\documentclass[11pt]{article}
\usepackage[preprint]{acl}
\usepackage{times}
\usepackage{latexsym}
\usepackage[T1]{fontenc}
\usepackage[utf8]{inputenc}
\usepackage{microtype}
\usepackage{inconsolata}
\usepackage{graphicx}
\graphicspath{{figures/}}

\usepackage{booktabs}
\usepackage{amsmath}
\usepackage{amssymb}
\usepackage{amsfonts} \usepackage{adjustbox} \usepackage{xurl}

\title{Empirical Margin Sensitivity: An Interventional Neuron-Level Probe of Post-Trained Medical LLMs}
\author{
  Aniruddh Shreyas \quad Aarnav Kembhavi \quad Minal Moharir \\
  RV College of Engineering, Bengaluru, India \\
  \texttt{\{aniruddhshreyas.cs23, aarnavkembhavi.cs23, minalmoharir\}@rvce.edu.in}
}

\begin{document}
\maketitle

\begin{abstract}
Medical post-training raises benchmark accuracy, but accuracy does not show which internal components a model's decisions depend on, or whether components that support one benchmark also support another. We apply the weight-space intervention paradigm of ROME and MEMIT at the granularity of one MLP down-projection column. Our measurement, Empirical Margin Sensitivity (EMS), replaces each coordinate magnitude of a column by its mean while preserving the column's sign pattern and $\ell_1$ scale, and records the resulting change in the logit margin between the correct option and its hardest-wrong competitor. This isolates the margin's dependence on a column's fine-grained coordinate structure while the column stays in place. We discover high-EMS columns (anchors) on MedQA in ten public checkpoints (four general-purpose bases and six medical post-trained variants) across the Gemma, Llama, Qwen, and AFM families, and measure the same columns on MedMCQA and PubMedQA. Mean-ablating each discovered anchor set lowers MedQA accuracy in all six post-trained models, by 6 to 28 points. In the two pure-SFT checkpoints, crushing anchors raises the PubMedQA margin on correctly answered items (negative EMS), and anchors with this property sit significantly deeper than the same model's other anchors; the four checkpoints trained with RL or mixed recipes show positive PubMedQA EMS and no significant depth shift. The depth association recurs in each of three held-out checkpoints. Because the checkpoints differ in data and pipeline as well as objective, we report these as per-checkpoint observations.
\end{abstract}

\section{Introduction}

Post-training strongly affects model behavior, and aggregate scores such as accuracy or pass@k say little about which internal components produce a gain or whether those components support related tasks \citep{wei2022cot, ouyang2022instructgpt, schulman2017proximal, xu2025reasoning, mirzadeh2025gsm}. Behavioral studies report that SFT tends to memorize while RL tends to generalize \citep{chu2025sftrl, guo2025deepseek}, but how this contrast appears at the level of individual neurons is unclear. In the medical checkpoints we study, the picture is more mixed than a memorize-versus-generalize split predicts.

Mechanistic interpretability has identified structured transformer and MLP computations 	\citep{elhage2021mathematical, geva2021transformer}, and weight-space interventions such as ROME and MEMIT have established that editing a small number of feed-forward parameters causally changes model behavior 	\citep{meng2022rome, meng2023memit}. Most neuron-level \emph{rankings}, however, remain correlational 	\citep{huang2025neurons, chen2025neurons}. We introduce a perturbation that removes a single down-projection column's learned structure while holding its sign and scale fixed, paired with a margin-based score, Empirical Margin Sensitivity (EMS), so that the resulting neuron rankings rest on direct interventions.

\subsection{Contributions}

This paper makes the following contributions:

\begin{itemize}\itemsep2pt
    \item \textbf{A column-level instance of an established causal paradigm.} We apply the weight-space intervention approach of ROME and MEMIT \citep{meng2022rome, meng2023memit} to individual down-projection columns, using a perturbation that preserves a column's sign and $\ell_1$ scale exactly while removing its coordinate structure, so the column's presence and its fine-grained tuning can be measured separately. Anchor selection on a disjoint split admits a bounded false-discovery rate, which we report. Mean-ablating each anchor set lowers MedQA accuracy in all six post-trained models.

    \item \textbf{Dependence and transfer can come apart.} AlphaMed-8B has the largest anchor dependence in the study ($-28.4$pp MedQA accuracy under ablation), yet its anchors have positive margin EMS on MedMCQA and PubMedQA, and ablating them leaves accuracy on those datasets unchanged ($0.0$, $+0.8$pp). In this checkpoint, strong dependence on the discovery benchmark coexists with positive transfer, so the two properties need separate measurement.

    \item \textbf{A depth signature of negative PubMedQA EMS in the pure-SFT checkpoints.} Some anchors have negative PubMedQA EMS even though their MedQA EMS is positive by construction: their fine-grained tuning raises the MedQA margin and lowers the PubMedQA margin on correctly answered items. In both pure-SFT checkpoints these anchors sit significantly deeper than the model's other anchors (Mann--Whitney U; MedGemma $+2.9$ layers, $p=0.002$; BioMed-L3 $+2.0$, $p=0.011$), with no significant shift in the four RL or mixed checkpoints.
\end{itemize}

\paragraph{Related work.}
Prior work identifies structured transformer subcomputations, interpretable MLP memories, and task-specific circuits 	\citep{elhage2021mathematical, geva2021transformer, elhage2022superposition, bricken2023monosemanticity, templeton2024scaling, wang2023ioi, conmy2023acdc}, but most neuron-level analyses remain correlational 	\citep{dai2022knowledge, wang2022skill, huang2025neurons, chen2025neurons, chen2025qrnca}. Other work examines how training-data statistics shape factual behavior and when parametric memory becomes unreliable 	\citep{elazar2023measuring, mallen2023trust}. CRANE 	\citep{le2026crane} validates language-specific neurons via targeted interventions but focuses on multilingual specialization; we instead compare neuron-level structure across post-trained checkpoints. Activation-magnitude and correlation-based rankings cannot distinguish a neuron that uniformly boosts all plausible medical terms from one that discriminates the correct answer from its strongest competitor; EMS scores the margin against the hardest-wrong distractor (Eqs.~\ref{eq:hardest}--\ref{eq:margin}) under a weight-space intervention. \citet{song2024taskneurons} identify task-specific neurons and validate them by deactivation; EMS differs in scoring a structure-removing weight edit against a competitor-specific margin, which lets it register neurons whose removal improves a held-out margin. This work joins two lines: ROME and MEMIT 	\citep{meng2022rome, meng2023memit} showed that targeted feed-forward edits causally control behavior, while activation patching, path patching, and DAS 	\citep{zhang2024activation, goldowsky2023localizing, wu2023boundlessdas} supply protocols for testing necessity. Neither supplies an intervention that leaves a neuron in place while removing its fine-grained coordinate structure, which is what makes tuning separately measurable from presence. The crush acts on the final column of each checkpoint. It therefore measures structure the checkpoint currently relies on; attributing that structure to a particular training stage requires the base-model comparisons of Section~\ref{sec:results}.

We build on behavioral evidence that SFT favors memorization while RL favors generalization 	\citep{chu2025sftrl, ziegler2019fine, stiennon2020learning, schulman2017proximal, guo2025deepseek}, and on evidence of structural weight-space adaptation, hierarchical knowledge traversal, and compositional reasoning gains under RL 	\citep{he2026structural, zhang2025hierarchical, cheng2025atomic}. We also engage the tension between claims that RL incentivizes correct reasoning 	\citep{wen2026rlvr} and evidence that reward optimization can induce memorization shortcuts 	\citep{yan2026spurious}, a question especially acute in medical LLMs 	\citep{wang2025medical, singhal2023large, lai2026medr1, pan2025medvlmr1}.

\section{Method}
\label{sec:method}

\subsection{Problem Setup}
We study multiple-choice question answering (MCQ) with a designated correct option. For each input question $x$, let $y^\star$ denote the correct answer token and let
\begin{equation}
\tilde{y} = \arg\max_{y \neq y^\star} \ell(y \mid x)
\label{eq:hardest}
\end{equation}
be the highest-logit incorrect option (the \emph{hardest wrong} distractor), where $\ell(\cdot\mid x)$ denotes the pre-softmax logit. We define the decision margin as
\begin{equation}
M(x) = \ell(y^\star \mid x) - \ell(\tilde{y} \mid x).
\label{eq:margin}
\end{equation}

A ``neuron'' here is a column of the MLP down-projection, the ``value'' vector of the key--value memory framework 	\citep{geva2021transformer, geva2022transformer}. In gated architectures (SwiGLU/GeGLU) this projection writes directly into the residual stream.

\subsection{Neuron Crush Intervention}
Consider an MLP column (neuron) $c$ with parameter vector $w_c \in \mathbb{R}^{d}$. We define a \emph{crushed} version
\begin{equation}
\hat{w}_c = \operatorname{sign}(w_c) \cdot \mathbb{E}[|w_c|],
\label{eq:crush}
\end{equation}
where $\mathbb{E}[|w_c|]$ is the mean absolute value over coordinates of $w_c$ and $\operatorname{sign}(\cdot)$ is elementwise. Unlike zero ablation, crushing preserves the sign pattern and the $\ell_1$ norm exactly ($\|\hat{w}_c\|_1 = \|w_c\|_1$): each coordinate magnitude is replaced by its mean, so the neuron's fine-grained tuning is removed while its sign and overall scale remain intact.

\subsection{Empirical Margin Sensitivity (EMS)}
For neuron $c$, we define Empirical Margin Sensitivity (EMS) as the expected margin drop under crushing:
\begin{equation}
\operatorname{EMS}(c) = \mathbb{E}_{x \sim \mathcal{D}}\left[M_{\text{base}}(x) - M_{\text{crush}(c)}(x)\right].
\end{equation}

Positive $\operatorname{EMS}(c)$ means the column's tuning supports the margin (crushing reduces it), negative means the tuning works against the margin (crushing increases it), and $\operatorname{EMS}(c)\approx 0$ means the margin is insensitive to it. EMS is a margin quantity: a nonzero value need not change any prediction, and the accuracy consequences of anchor removal are measured separately (Section~\ref{sec:results_patching}).

\subsection{Anchor Discovery Pipeline}
\begin{figure*}[!ht]
\centering
\includegraphics[width=1\textwidth]{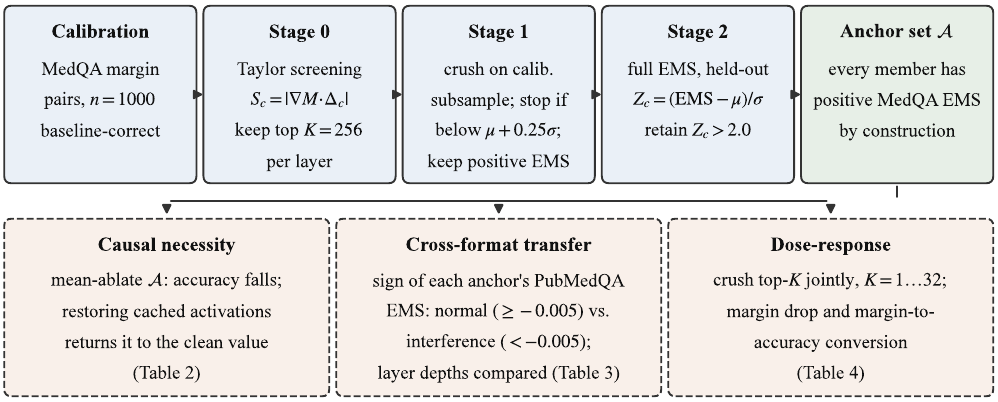}
\caption{The three-stage anchor discovery pipeline (top) and the downstream analyses it feeds (bottom). Stage~2 validation is one-tailed, so every member of the anchor set $\mathcal{A}$ has positive MedQA EMS by construction. The downstream measurements do not feed back into discovery: interference is read off the sign of each anchor's PubMedQA EMS, which the discovery criterion leaves unconstrained.}
\label{fig:pipeline}
\end{figure*}

To isolate the margin's dependence on a neuron's fine-grained tuning rather than on its presence, we use a staged pipeline (Figure~\ref{fig:pipeline}).

\paragraph{Stage 0: Taylor screening.}
For each sample, we compute the gradient of the margin with respect to each candidate column and accumulate over the calibration set. Let
\begin{equation}
\Delta_c = \hat{w}_c - w_c,
\end{equation}
be the crush perturbation, and let $\nabla_{w_c} M$ denote the gradient of the decision margin $M(x)$ (Eq.~\ref{eq:margin}) with respect to neuron $c$'s down-projection column $w_c$, all other parameters fixed. We define the first-order screening score
\begin{equation}
S_c = \left|\nabla_{w_c} M \cdot \Delta_c\right|,
\end{equation}
the absolute first-order margin change predicted for the crush perturbation. We rank columns by $S_c$ and retain the top candidates per layer. This inexpensive stage filters the search space using directional sensitivity.

\paragraph{Stage 1: EMS with early stopping.}
Each candidate is evaluated on calibration examples using direct crush interventions. We stop early if its cumulative mean drop falls below
\begin{equation}
\mu_{\text{rand}} + 0.25\,\sigma_{\text{rand}},
\end{equation}
where $\mu_{\text{rand}}$ and $\sigma_{\text{rand}}$ are random-baseline statistics. Candidates with positive mean drop are retained.

\paragraph{Stage 2: full EMS and Z-score validation.}
Stage 2 uses data disjoint from calibration. For retained candidates, we compute full EMS on a held-out set and compare each candidate with a random-column EMS distribution using
\begin{equation}
Z_c = \frac{\operatorname{EMS}(c) - \mu_{\text{rand}}}{\sigma_{\text{rand}}}.
\end{equation}

A neuron is declared an anchor when its standardized score exceeds $2.0$ (one-tailed, nominal $p \approx 0.023$) against the single-column random null. Both thresholds were fixed on a disjoint validation split before any experiments in Section~\ref{sec:results} and are used throughout without further tuning (appendix).

\paragraph{Operational definition.} An \emph{anchor} is a column that passes this criterion and nothing more: the term does not imply semantic content, uniqueness, or membership in a circuit.

\paragraph{False-discovery control.} The threshold is applied per candidate, so some anchors are expected to be null columns. Because Stage~2 uses data disjoint from the Stage~0--1 calibration set, each Stage-2 test is valid at its nominal level, and with $N$ Stage-2 candidates the expected number of null columns passing is at most $0.023N$. With $N$ between 451 and 512 per fine-tuned model (appendix), this bounds the expected false-discovery proportion of each anchor set at 18\% (MedGemma, BioMed-L3, AlphaMed-8B, Meditron), 25\% (AlphaMed-7B), and 27\% (AFM-OpenMed). The bound assumes every candidate is null and is therefore conservative. Null columns dilute set-level averages toward the random baseline, so the anchor-set effects below are, if anything, attenuated by them.

\subsection{Random Baseline Estimation}
The random baseline is a single-column null: one randomly chosen down-projection column is crushed at a time and its margin drop recorded, pooled over random column sets size-matched to each model's anchor count to give $(\mu_{\text{rand}}, \sigma_{\text{rand}})$. Anchor drops in Table~\ref{tab:anchor_vs_random_combined} are computed the same way over the anchor set, so size matching refers to the number of columns entering the average; columns are crushed one at a time. Random columns are drawn uniformly over all layers and are matched to the anchor set in count only, so the baseline does not control for layer or weight norm. Standard deviations there are taken across the $n$ evaluation questions, and Cohen's $d$ uses those per-question standard deviations (appendix).

\paragraph{Relation to a Taylor expansion.}
A second-order expansion of EMS separates a first-order sensitivity term from a curvature correction, and the crush perturbation satisfies exactly $\|\Delta_c\|_2=\sqrt{d}\,\sigma(|w_c|)$, bounding that correction under local smoothness (appendix). Because this ties the perturbation size to the column's weight dispersion, we test whether EMS reduces to that dispersion: across Stage-2 candidates the rank correlation between EMS and $\sigma(|w_c|)$ is at most $0.24$ in every model (appendix), so anchor rankings are not recoverable from weight dispersion. The first-order term motivates only the Stage 0 screening score. The EMS values used in Stages 1--2 are finite-difference measurements with the crushed weights in place, so they include curvature and all higher-order terms. We estimate EMS by the sample mean over $n$ evaluation questions, unbiased under i.i.d.\ sampling.

\section{Experimental Setup}

\subsection{Models and Architectures}
We evaluate ten models (four pretrained bases, six fine-tuned variants) across the Gemma, Llama, Qwen, and AFM families: Gemma-4B/MedGemma (SFT); Llama-3-8B/BioMed-L3 (SFT) and AlphaMed-8B (RL); Qwen-2.5-7B/Meditron (CPT+SFT) and AlphaMed-7B (RL); AFM-4.5B/AFM-OpenMed (SFT+DPO+RL). Layer counts, widths, and model references are in the appendix. Each family pairs a base with post-trained variants spanning supervised fine-tuning (SFT), continual pretraining then SFT (CPT+SFT), reinforcement learning (RL), and a multi-stage SFT, preference and RL recipe. Within the Llama and Qwen families, variants share architecture and scale but start from different released checkpoints and data, so within-family contrasts compare complete training recipes. Recipe labels follow each model's public card. The design is observational: each recipe category contains two checkpoints, and checkpoints differ in training data, data scale, curriculum, and optimization as well as objective, so no contrast here isolates the effect of the objective. The search space---$663{,}552$ candidate columns for AFM-4.5B---motivates the staged pipeline.

\subsection{Datasets and Evaluation Protocol}

We use three medical QA benchmarks: \textbf{MedQA} 	\citep{jin2021disease} (4-option MCQ) for anchor discovery, \textbf{MedMCQA} 	\citep{pal2022medmcqa} (4-option MCQ) as a held-out benchmark close to MedQA in format, and \textbf{PubMedQA} 	\citep{jin2019pubmedqa} as a held-out benchmark further from it. PubMedQA is evaluated as a standardized 4-option MCQ (A: yes, B: no, C: maybe, D: contradictory evidence), matching the answer format of MedQA/MedMCQA but differing in context length and option naturalness: C and D are fixed templated strings. PubMedQA also differs from MedQA in source, context structure, question style, and domain distribution. We therefore refer to the two held-out benchmarks by name and attribute no PubMedQA effect to format specifically; the appendix rules out the templated option D as the source of the sign. The margin-based EMS analyses use items filtered to baseline-correct questions, where the margin is positive and a drop is interpretable; activation-patching accuracies (Table~\ref{tab:activation_patching}) use the full evaluation set, so Clean is each model's unfiltered baseline and $\Delta$Acc is a paired within-question change.

The hardest-wrong option $\tilde{y}$ is computed once per question on the unmodified model and held fixed across interventions. $M$ is therefore a fixed-competitor margin. It equals the post-intervention decision margin whenever the identity of the strongest wrong option is unchanged, and upper-bounds it otherwise. Positive EMS is therefore a lower bound on the drop in decision margin, while negative EMS is an upper bound on its rise; the negative PubMedQA values below should be read with this direction in mind. Fixing $\tilde{y}$ also keeps $M$ differentiable in $w_c$, which Stage~0 requires (appendix). Welch's $t$-test statistics for every model--dataset pair are reported in the appendix.

\subsection{Implementation Details and Sample Sizes}
\textbf{Compute.} Single NVIDIA H100 (80GB), Linux, PyTorch 2.x, bfloat16.

\textbf{Sample sizes and hyperparameters.} EMS calibration uses 1000 MedQA margin pairs; Stage 2 validation, activation patching, and the K-sweep use 500, 240$\times$5, and 300 samples respectively; anchor-vs-random uses 100 trials $\times$ 5 seeds with random groups size-matched to each model's anchor count (full details in the appendix). Stage 0 retains $K=256$ per layer; Stage 2 uses $Z_c>2.0$ against a matched random-neuron baseline.

Code, configuration, and instructions to reproduce all reported results are available at \url{https://github.com/aarnav-kembhavi/MEDNSQ-CIRCUITS}.

\section{Results}
\label{sec:results}

For readability we label the six fine-tuned checkpoints by recipe: pure SFT (MedGemma, BioMed-L3), pure RL (AlphaMed-7B, AlphaMed-8B), and mixed (Meditron, AFM-OpenMed). With two checkpoints per label, every recipe-level statement below describes these checkpoints and is not an estimate of the effect of an objective. The clearest separation is the sign of PubMedQA EMS (negative for both pure-SFT checkpoints, positive for the other four) and the depth of negative-EMS anchors, which shifts significantly only in the two pure-SFT checkpoints.

\subsection{Anchor Discovery in Base and Post-Trained Checkpoints}
\label{sec:results_anchor_discovery}

Pretrained models yield small anchor sets (Gemma-4B: 9, Qwen-2.5-7B: 12, AFM-4.5B: 19, Llama-3-8B: 31), and every post-trained variant yields more at the same threshold (MedGemma: 66, BioMed-L3: 64, AlphaMed-8B: 64, Meditron: 64, AlphaMed-7B: 43, AFM-OpenMed: 38). Base-model anchor sets are weakly identified: on fresh evaluation data their effect over count-matched random columns is small in every base ($|d|\le0.26$ on MedQA; appendix), whereas post-trained anchor sets separate clearly (Table~\ref{tab:anchor_vs_random_combined}). For this reason we report overlap between base and post-trained anchor sets only descriptively in the appendix and draw no conclusion from it.

\subsection{Anchor Effects on Discovery and Held-Out Benchmarks}
\label{sec:results_anchor_vs_random}

\begin{figure*}[t]
\centering
\includegraphics[width=\textwidth]{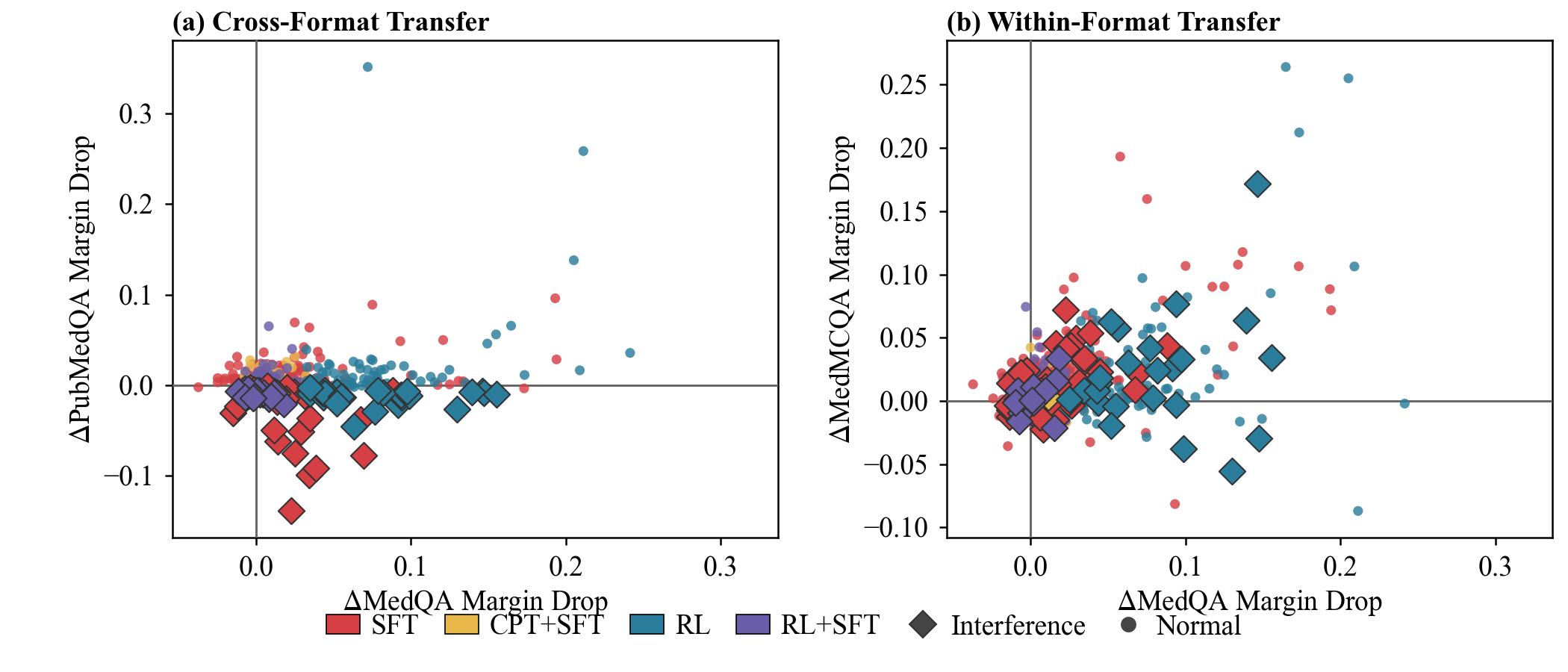}
\caption{Per-anchor transfer geometry across post-trained checkpoints. Each point is a discovered anchor; diamonds are negative-EMS anchors (PubMedQA EMS $<-0.005$), circles the remaining anchors. (a) MedQA vs PubMedQA: both pure-SFT checkpoints concentrate in the negative-EMS region, while RL and mixed-recipe anchors lie mostly in the positive region. This is a checkpoint-level pattern; individual anchors in every checkpoint can have negative EMS. (b) MedQA vs MedMCQA. Per-model interference rates are in Table~\ref{tab:interference_layer_shift}.}
\label{fig:transfer_geometry}
\end{figure*}

We compare anchor-set crush effects against count-matched random columns on $n=500$ fresh questions per benchmark (Table~\ref{tab:anchor_vs_random_combined}). Random drops stay near zero while anchors show consistent directional effects. Both pure-SFT checkpoints show large positive MedQA and MedMCQA effects and negative PubMedQA effects of similar standardized size ($d\approx-1.0$; raw drops $-0.018$ and $-0.026$). The four RL or mixed checkpoints show positive effects on all three benchmarks ($d$ from $+0.40$ to $+7.90$; all $p<0.001$). Figure~\ref{fig:transfer_geometry} shows the per-anchor geometry; in margin terms all six fine-tuned checkpoints transfer to MedMCQA. Neither SFT checkpoint's base shows a negative PubMedQA anchor effect (Gemma-4B $-0.0003$, at the noise floor; Llama-3-8B $+0.0015$; appendix), against $-0.0180$ for MedGemma and $-0.0263$ for BioMed-L3. Because base anchor sets are weakly identified and the crush acts on final weights, this comparison locates the negative sign in the post-trained checkpoints without attributing it to a training stage. Among our ten checkpoints, negative PubMedQA EMS at the anchor-set level occurs only in the two pure-SFT models. The sign is unchanged when the hardest-wrong distractor is restricted to the native yes/no/maybe options (appendix).

\begin{table*}[t]
\centering
\caption{Anchor vs.\ random drops, reported as mean per-neuron margin drop over the anchor set against count-matched random neurons; mean $\pm$ per-question std and Cohen's $d$ over $n=500$ evaluation questions disjoint from both the calibration and Stage-2 selection sets. Anchors are discovered on MedQA, so the MedMCQA and PubMedQA rows are held out from discovery. Negative PubMedQA $d$ occurs only for the two pure-SFT checkpoints. Details in the appendix.}
\label{tab:anchor_vs_random_combined}

\small
\setlength{\tabcolsep}{3pt}
\renewcommand{\arraystretch}{1.05}
\begin{adjustbox}{max width=\linewidth}\begin{tabular}{l cc cc cc}
\toprule
& \multicolumn{2}{c}{Pure SFT} & \multicolumn{2}{c}{Pure RL} & \multicolumn{2}{c}{Mixed} \\
\cmidrule(lr){2-3}\cmidrule(lr){4-5}\cmidrule(lr){6-7}
& MedGemma & BioMed-L3 & AlphaMed-8B & AlphaMed-7B & Meditron & AFM-OpenMed \\
\midrule
\multicolumn{7}{l}{\textit{MedQA} (discovery format)} \\
\quad Anchor & $+0.0335\pm0.0232$ & $+0.1042\pm0.1034$ & $+0.1797\pm0.1994$ & $+0.0837\pm0.0140$ & $+0.0455\pm0.0099$ & $+0.0320\pm0.0092$ \\
\quad Random & $+0.0007\pm0.0176$ & $+0.0156\pm0.0113$ & $+0.0071\pm0.0021$ & $+0.0091\pm0.0012$ & $-0.0012\pm0.0048$ & $+0.0013\pm0.0048$ \\
\quad $d$    & $+1.59$ & $+1.20$ & $+1.22$ & $+7.51$ & $+6.00$ & $+4.18$ \\
\midrule
\multicolumn{7}{l}{\textit{MedMCQA} (held out, MCQ-matched)} \\
\quad Anchor & $+0.0960\pm0.0304$ & $+0.0429\pm0.0792$ & $+0.0320\pm0.0957$ & $+0.0415\pm0.0072$ & $+0.0110\pm0.0098$ & $+0.0710\pm0.0152$ \\
\quad Random & $+0.0121\pm0.0151$ & $+0.0092\pm0.0117$ & $+0.0036\pm0.0011$ & $+0.0051\pm0.0086$ & $+0.0001\pm0.0027$ & $+0.0067\pm0.0077$ \\
\quad $d$    & $+3.50$ & $+0.59$ & $+0.42$ & $+4.59$ & $+1.52$ & $+5.34$ \\
\midrule
\multicolumn{7}{l}{\textit{PubMedQA} (held out, shifted)} \\
\quad Anchor & $-0.0180\pm0.0300$ & $-0.0263\pm0.0407$ & $+0.0308\pm0.0955$ & $+0.0105\pm0.0012$ & $+0.0475\pm0.0083$ & $+0.0195\pm0.0154$ \\
\quad Random & $+0.0053\pm0.0125$ & $+0.0038\pm0.0101$ & $+0.0041\pm0.0007$ & $+0.0047\pm0.0003$ & $-0.0002\pm0.0020$ & $-0.0005\pm0.0054$ \\
\quad $d$    & $-1.01$ & $-1.02$ & $+0.40$ & $+6.63$ & $+7.90$ & $+1.73$ \\
\bottomrule
\end{tabular}\end{adjustbox}
\end{table*}

\subsection{Activation Patching: Necessity under Mean Ablation}
\label{sec:results_patching}

\paragraph{Patching intervention.} Activation patching intervenes on the anchor neurons' post-activation outputs: the corrupted run replaces each anchor neuron's activation with its dataset-mean activation (mean ablation), the restored run reinstates the clean cached activation, and accuracy is measured under each on the full, unfiltered evaluation set. This is a second intervention, distinct from the weight-space crush of Eq.~\eqref{eq:crush}: the crush ranks neurons by margin sensitivity, and mean ablation tests whether the resulting set is necessary for accuracy. Mean ablation can place activations off their natural distribution, so these effects describe necessity under this intervention specifically, and no random-set patching control is included.

\begin{table}[!ht]
\centering
\caption{Activation patching results on the full evaluation set. $\Delta$Acc is corrupted-minus-clean accuracy change after mean-ablating the anchor set: negative indicates dependence; $^*$ marks an accuracy gain. Rest.\ reinstates cached clean activations and returns accuracy to the clean value in every cell, confirming that the change is due to the intervention; the dependence result rests on the Corr.\ column.}
\label{tab:activation_patching}

\small
\setlength{\tabcolsep}{4pt}
\renewcommand{\arraystretch}{0.9}
\begin{adjustbox}{max width=\linewidth}\begin{tabular}{l l c c c c}
\toprule
Model & Data & Clean & Corr. & Rest. & $\Delta$Acc \\
\midrule

MedGemma    & MedQA & 62.1 & 47.9 & 62.1 & $-14.2\pm1.2$ \\
            & MedMCQA & 65.4 & 66.7 & 65.4 & $+1.3\pm0.5^*$ \\
            & PubMedQA  & 76.2 & 75.8 & 76.2 & $-0.4\pm1.2$ \\
\midrule

BioMed-L3   & MedQA & 73.3 & 60.8 & 73.3 & $-12.5\pm1.9$ \\
            & MedMCQA & 80.4 & 76.7 & 80.4 & $-3.7\pm2.3$ \\
            & PubMedQA  & 80.0 & 76.7 & 80.0 & $-3.3\pm2.0$ \\
\midrule

AlphaMed-8B & MedQA & 71.3 & 42.9 & 71.3 & $-28.4\pm4.4$ \\
            & MedMCQA & 85.0 & 85.0 & 85.0 & $+0.0\pm1.9$ \\
           & PubMedQA  & 77.5 & 78.3 & 77.5 & $+0.8\pm2.4$ \\
\midrule

Meditron    & MedQA & 60.4 & 47.1 & 60.4 & $-13.3\pm1.2$ \\
            & MedMCQA & 70.0 & 68.3 & 70.0 & $-1.7\pm2.4$ \\
            & PubMedQA  & 78.3 & 77.1 & 78.3 & $-1.2\pm1.2$ \\
\midrule

AlphaMed-7B & MedQA & 69.6 & 57.5 & 69.6 & $-12.1\pm1.2$ \\
            & MedMCQA & 70.4 & 65.4 & 70.4 & $-5.0\pm2.0$ \\
            & PubMedQA  & 75.4 & 75.0 & 75.4 & $-0.4\pm1.2$ \\
\midrule

AFM-OpenMed & MedQA & 50.8 & 44.6 & 50.8 & $-6.2\pm1.4$ \\
            & MedMCQA & 58.8 & 60.4 & 58.8 & $+1.6\pm0.6^*$ \\
            & PubMedQA  & 74.2 & 73.8 & 74.2 & $-0.4\pm0.6$ \\

\bottomrule
\end{tabular}\end{adjustbox}
\end{table}

Because anchors are discovered only on MedQA, the MedMCQA and PubMedQA rows are held out from discovery. MedQA shows the strongest dependence: $-12$ to $-14$pp for four models, $-6.2$pp for AFM-OpenMed, and $-28.4$pp for AlphaMed-8B, the largest effect in the study.

\paragraph{Margin effects and accuracy effects.} The PubMedQA rows bear directly on how negative EMS should be read. If the pure-SFT anchors' tuning lowered PubMedQA accuracy, removing the set would raise it. It does not: mean-ablating the full anchor set changes PubMedQA accuracy by $-0.4$pp in MedGemma, within the $1$pp resolution of the evaluation set (appendix), and by $-3.3$pp in BioMed-L3. Negative PubMedQA EMS is therefore a margin effect on correctly answered items, and we use it only in that sense. Two differences between the measurements may account for the gap: mean ablation is a different intervention from the crush, and the ablated set mixes negative-EMS anchors with positive-EMS ones, whose removal can cost accuracy. Whether ablating the negative-EMS subset alone changes PubMedQA accuracy is not tested here.

\subsection{Layer Distribution of Negative-EMS Anchors}
\label{sec:interference}

We call an anchor \emph{interfering} when its PubMedQA EMS is below $\tau=-0.005$, and \emph{normal} otherwise. The label is operational and refers only to the margin effect defined above. $\tau$ was set from random-baseline noise before the analysis, and the prevalence ordering is unchanged for $\tau$ from $-0.001$ to $-0.010$ (appendix). Table~\ref{tab:interference_layer_shift} reports counts, percentages, and mean layer indices.

\begin{figure*}[!t]
\centering
\includegraphics[width=1\textwidth]{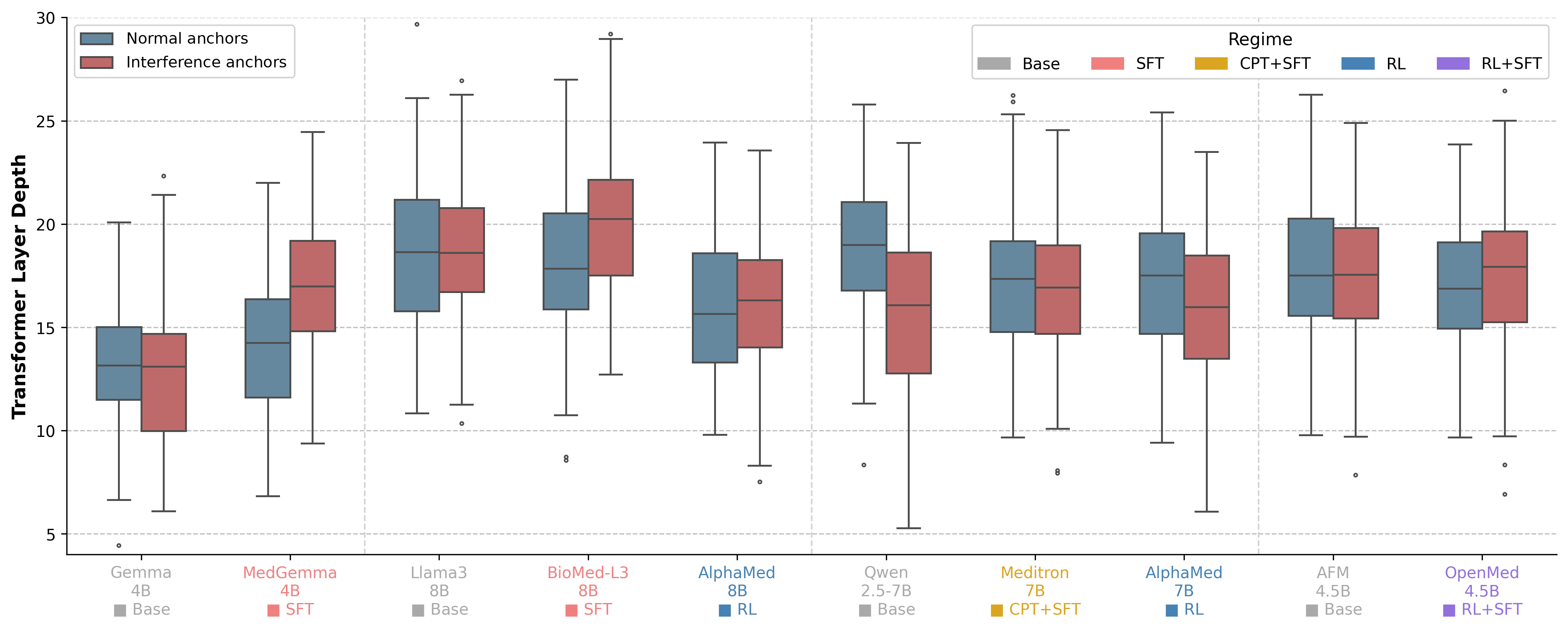}
\caption{Transformer layer depth distribution for interfering (red) and normal (blue) anchors in each checkpoint. Interfering anchors are those with PubMedQA EMS $<-0.005$.}
\label{fig:interference_depth}
\end{figure*}

\begin{table}[t]
\centering
\caption{Layer depth of interfering (PubMedQA EMS $<-0.005$) and normal anchors. Shift = interfering mean layer $-$ normal mean layer. One-sided Mann--Whitney U $p$; across the six fine-tuned checkpoints only MedGemma and BioMed-L3 reach significance, and both survive Benjamini--Hochberg correction (appendix).}
\label{tab:interference_layer_shift}

\small
\setlength{\tabcolsep}{4pt}
\renewcommand{\arraystretch}{0.9}
\begin{adjustbox}{max width=\linewidth}\begin{tabular}{l c c c c c}
\toprule
Model & Interf.\% & Interf. Mean & Norm. Mean & Shift & $p$ \\
\midrule

Gemma-4B    & 66.7 & 12.8 & 13.6 & $-$0.8 & 0.81 \\
MedGemma    & 34.8 & 16.8 & 13.9 & +2.9 & 0.002 \\
\midrule

Llama-3-8B  & 51.6 & 19.0 & 18.9 & +0.0 & 0.50 \\
BioMed-L3   & 48.4 & 20.0 & 18.0 & +2.0 & 0.011 \\
AlphaMed-8B & 37.5 & 16.0 & 15.1 & +0.9 & 0.19 \\
\midrule

Qwen-2.5-7B & 33.3 & 15.4 & 18.3 & $-$2.8 & 0.99 \\
Meditron    & 10.9 & 17.0 & 17.2 & $-$0.2 & 0.52 \\
AlphaMed-7B & 30.2 & 16.3 & 16.3 & +0.0 & 0.72 \\

\midrule

AFM-4.5B    & 31.6 & 17.3 & 17.2 & +0.1 & 0.46 \\
AFM-OpenMed & 21.1 & 17.2 & 17.2 & +0.0 & 0.62 \\

\bottomrule
\end{tabular}\end{adjustbox}
\end{table}

Each depth test compares a checkpoint's interfering anchors with its own normal anchors, so it is a within-model comparison and needs no cross-model pooling (Figure~\ref{fig:interference_depth}, Table~\ref{tab:interference_layer_shift}). Tests are one-sided in the deeper direction. In both pure-SFT checkpoints interfering anchors sit deeper than normal anchors: $+2.9$ layers in MedGemma ($p=0.002$) and $+2.0$ in BioMed-L3 ($p=0.011$). Across the six fine-tuned checkpoints both survive Benjamini--Hochberg correction ($q=0.012$, $0.033$); under the more conservative Holm procedure BioMed-L3 is borderline ($p_{\mathrm{Holm}}=0.055$; appendix). None of the four RL or mixed checkpoints shows a significant shift (all $p\ge0.19$). The two mixed-recipe checkpoints have the lowest interfering-anchor prevalence (10.9\% and 21.1\%).

Prevalence alone does not separate base from post-trained checkpoints: Llama-3-8B already shows 51.6\% against BioMed-L3's 48.4\% (Gemma-4B's 66.7\% sits on drops of order $10^{-4}$ and is not interpreted; appendix). In the SFT checkpoints, the measurable difference from the base is the \emph{depth} of interfering anchors.

\paragraph{Held-out checkpoints.} We ran discovery on three further checkpoints that played no role in developing EMS or setting thresholds (Med42-8B \citep{med42v2_2024}, BioMistral-7B \citep{biomistral2024}, and AlphaMed-3B, the base-rl checkpoint from \citet{alphamed2025}), binarizing depth at the median layer (17) of the six-model corpus and keeping $\tau=-0.005$, without refitting. Interfering anchors are over-represented in deep layers in each of the three (odds ratios $3.05$, $2.08$, $6.08$). Three of three in the same direction gives an exact one-sided sign test of $p=0.125$ at the model level; anchor-level pooling (Fisher's exact odds ratio $3.46$; appendix) treats anchors within a model as independent and so does not measure replication across models. The held-out set spans SFT+DPO, CPT and RL recipes and its strongest association is in the RL checkpoint, so it supports a general depth--interference association; the SFT-specific statement rests on the within-model shifts in Table~\ref{tab:interference_layer_shift}.

\subsection{Group Ablation Dose-Response}
\label{sec:results_group_ablation}

To relate margin loss to accuracy loss, we crush the top-$K$ anchors jointly for $K\in\{1,2,4,8,16,32\}$ and record the margin drop and accuracy change. Raw margins depend on each model's logit scale, so cross-model comparisons use the margin drop normalized by the model's clean mean margin.

\begin{table}[t]
\centering
\caption{Group ablation dose-response at representative $K$; the full sweep ($K\in\{1,2,4,8,16,32\}$) is plotted in Figure~\ref{fig:three_panel}. Margin rows are raw logit-margin drops on baseline-correct items and are comparable only within a model; $\Delta$Acc rows use the full evaluation set. Norm.\ Drop is the $K{=}32$ margin drop as a fraction of the model's clean mean margin. Models abbreviated: MedG, BioL3, Alpha8B, Alpha7B, Medit.\ (Meditron), Open (AFM-OpenMed).}
\label{tab:k_sweep_full}

\small
\setlength{\tabcolsep}{3pt}
\renewcommand{\arraystretch}{0.85}
\begin{adjustbox}{max width=\linewidth}\begin{tabular}{c l cccccc}
\toprule
K & Metric & MedG & BioL3 & Alpha8B & Alpha7B & Medit. & Open \\
\midrule

1  & Margin       & 0.036 & 0.186 & 0.763 & 0.595 & 0.060 & 0.042 \\
   & $\Delta$Acc  & $-$1.67\% & $-$1.00\% & $-$3.00\% & +0.33\% & +1.33\% & 0.00\% \\

\midrule
8  & Margin       & 0.432 & 0.913 & 1.678 & 1.402 & 0.324 & 0.213 \\
   & $\Delta$Acc  & $-$2.00\% & $-$4.00\% & $-$4.00\% & $-$1.33\% & $-$1.67\% & $-$1.00\% \\

\midrule
32 & Margin       & 1.690 & 1.356 & 3.198 & 2.697 & 0.644 & 0.530 \\
   & $\Delta$Acc  & $-$5.67\% & $-$7.00\% & $-$8.67\% & $-$8.00\% & $-$6.67\% & $-$6.00\% \\
   & Norm. Drop   & 9.19\% & 11.29\% & 11.60\% & 12.77\% & 13.13\% & 9.93\% \\

\bottomrule
\end{tabular}\end{adjustbox}

\end{table}

\begin{figure*}[t]
\centering
\includegraphics[width=1\textwidth]{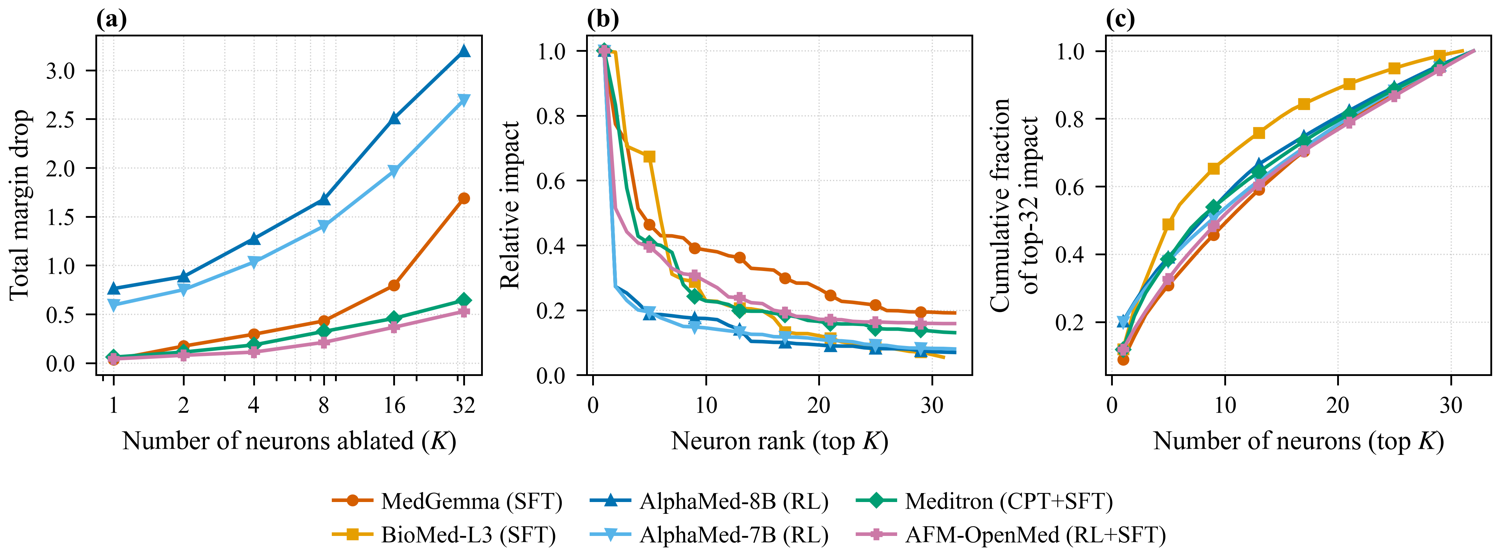}
\caption{
Three-panel analysis of anchor structure.
(a) Margin-dose response vs.\ anchor count $K$.
(b) Decay of per-neuron contribution with rank.
(c) Cumulative top-$K$ margin drop.
}
\label{fig:three_panel}
\end{figure*}

Margin drop grows monotonically with $K$ in every model (Figure~\ref{fig:three_panel}a), and per-neuron contribution decays with rank (Figure~\ref{fig:three_panel}b). At $K{=}32$, crushing removes 9 to 13\% of the clean margin and costs 5.7 to 8.7 points of accuracy (Table~\ref{tab:k_sweep_full}). The ratio of accuracy loss to normalized margin drop lies between $0.51$ and $0.75$ in all six checkpoints, so the checkpoints do not separate by recipe on how margin loss converts into errors.

\section{Discussion}
\label{sec:discussion}

Throughout, \textit{interfering} refers to anchors with PubMedQA EMS below $\tau$. It is an operational descriptor of a margin effect and makes no claim about mechanism or accuracy. As an exploratory check we also ablate attention heads jointly with anchors (appendix).

\paragraph{Dependence and transfer.} The most anchor-dependent checkpoint in the study is RL-trained (AlphaMed-8B, $-28.4$pp MedQA accuracy under mean ablation), and its anchors still have positive EMS on both held-out benchmarks, with no accuracy cost there when ablated. In this checkpoint, heavy dependence on a small anchor set coexists with positive transfer, which is the opposite of the brittle, non-transferring shortcuts sometimes attributed to reward optimization \citep{yan2026spurious}. This is one checkpoint, so it shows that the two properties can come apart, and it does not show how often they do. Behavioral accounts often report them together \citep{wen2026rlvr}; measuring them separately, as here, is what makes the distinction visible.

\paragraph{Within-architecture comparison.} Meditron and AlphaMed-7B share the Qwen-2.5-7B architecture and scale, which holds both fixed across the two recipes while data and pipeline still differ. Meditron has the lowest interfering-anchor prevalence in the study (10.9\%, against 30.2\% for AlphaMed-7B), and neither shows a depth shift \citep{wang2022skill, cheng2025atomic, yan2026spurious}.

\paragraph{What a controlled study would add.} The recipe contrasts here are observational. Training SFT and RL variants from one base on shared data would isolate the objective; crushing only the base-to-fine-tuned weight change would attribute structure to post-training; and ablating the interfering subset alone would test whether negative PubMedQA EMS carries an accuracy cost. EMS applies unchanged in each of these designs.

\section{Conclusion}
Taking the weight-space intervention paradigm down to individual down-projection columns, and checking the result with activation patching, gives a margin-level view of which columns a post-trained model's decisions depend on. Anchor sets discovered on MedQA are necessary for MedQA accuracy under mean ablation in all six post-trained checkpoints. In the two pure-SFT checkpoints, a subset of anchors lowers the PubMedQA margin and sits deeper than the remaining anchors; the four RL and mixed-recipe checkpoints show positive PubMedQA EMS and no depth shift, and a deep-layer association recurs in three held-out checkpoints. These are observations on public checkpoints that differ in more than objective. EMS measures margin dependence and transfer separately, and the same procedure can be applied directly to controlled training comparisons.

\section*{Ethical Statement}

Medical language models are evaluated mainly by benchmark accuracy, which does not show what a model's decisions depend on internally. We release only analysis code on public checkpoints and present EMS as an exploratory research measurement. It complements standard accuracy evaluation and does not replace or certify it, and it should not inform clinical-use decisions. In particular, absence of a detected interference signature is not evidence of clinical safety, and EMS is studied here only on English multiple-choice benchmarks.

\bibliography{main}
\appendix

\section{Scope and Notation} \label{app:notation}

\begin{table}[t]
\centering
\footnotesize
\setlength{\tabcolsep}{2.5pt}
\renewcommand{\arraystretch}{0.95}
\caption{Model architectures and training regimes. Base/fine-tuned families include Gemma~3 \citep{gemma3_2025}, MedGemma \citep{medgemma2025}, Llama 3 \citep{llama32024}, Bio-Medical-Llama-3 (BioMed-L3) \citep{ContactDoctor_Bio-Medical-Llama-3-8B}, AlphaMed \citep{alphamed2025}, Qwen2.5 \citep{qwen25_2025}, Meditron-Qwen (Meditron) \citep{meditron_qwen}, AFM/OpenMed \citep{arcee2025afm45b,afm_openmed}, Med42-v2 \citep{med42v2_2024}, and BioMistral \citep{biomistral2024}.}
\label{tab:models}

\begin{adjustbox}{max width=\linewidth}\begin{tabular}{l l c c c c}
\toprule
Model & Train. & Layers & Hidden & FFN & Hd \\
\midrule
Gemma-4B        & Base    & 34 & 2560 & 10240 & 8  \\
MedGemma        & SFT     & 34 & 2560 & 10240 & 8  \\
\midrule
Llama-3-8B      & Base    & 32 & 4096 & 14336 & 32 \\
BioMed-L3       & SFT     & 32 & 4096 & 14336 & 32 \\
AlphaMed-8B     & RL      & 32 & 4096 & 14336 & 32 \\
\midrule
Qwen-2.5-7B     & Base    & 28 & 3584 & 18944 & 28 \\
Meditron        & CPT+SFT & 28 & 3584 & 18944 & 28 \\
AlphaMed-7B     & RL      & 28 & 3584 & 18944 & 28 \\
\midrule
AFM-4.5B        & Base    & 36 & 2560 & 18432 & 20 \\
AFM-OpenMed     & SFT+DPO+RL & 36 & 2560 & 18432 & 20 \\
\midrule
\multicolumn{6}{l}{\textit{Held-out check (Sec.~\ref{app:extended_quantitative})}} \\
Med42-8B     & SFT+DPO & 32 & 4096 & 14336 & 32 \\
BioMistral-7B & CPT  & 32 & 4096 & 14336 & 32 \\
AlphaMed-3B & RL & 36 & 2048 & 11008 & 16 \\
\bottomrule
\end{tabular}\end{adjustbox}
\end{table}

This appendix provides methodological detail supporting the main-paper analysis. Model names follow the short forms of Table~\ref{tab:models} (Gemma-4B, MedGemma, Llama-3-8B, BioMed-L3, AlphaMed-8B, Qwen-2.5-7B, Meditron, AlphaMed-7B, AFM-4.5B, AFM-OpenMed).

We fix notation once and use it throughout. Departures from the main paper's shorthand are noted explicitly.

\begin{itemize}\itemsep2pt
  \item $x$: an evaluation question; $y^\star$: its correct option token; $\tilde{y}$: its hardest-wrong option token.
  \item $\ell(y\mid x)$: pre-softmax logit of option $y$; $M(x)$: the decision margin (Eq.~\ref{eq:margin}).
  \item $c$: an MLP down-projection column (a \emph{neuron}); $w_c\in\mathbb{R}^{d}$: its parameter vector; $d$: the model's hidden width (the column's length).
  \item $L$: number of transformer layers; $d_{\mathrm{ff}}$: feed-forward width; the candidate space has $L\times d_{\mathrm{ff}}$ columns.
  \item $\hat{w}_c$: the crushed column; $\Delta_c=\hat{w}_c-w_c$: the crush perturbation.
  \item $m_c=\mathbb{E}[|w_c|]=\tfrac{1}{d}\|w_c\|_1$: mean coordinate magnitude; $\sigma(|w_c|)$: standard deviation of the coordinate magnitudes.
  \item $\mathcal{A}$: the validated anchor set for a model.
\end{itemize}

\paragraph{The random baseline, stated once.}
All random baselines in this paper are \emph{single-column} nulls
$(\mu^{\mathrm{col}},\sigma^{\mathrm{col}})$: one randomly chosen down-projection
column is crushed at a time and the resulting margin drop recorded. This single
null serves three purposes: the Stage-1 early-stopping rule, the Stage-2
standardized score $Z_c$, and the ``Random Drop'' column of main-paper Table~1.
In the last case each anchor's drop is likewise measured one column at a time
and averaged over the anchor set, and the random comparison is the same quantity
averaged over an equally sized set of randomly drawn columns. ``Size-matched''
therefore refers to the number of columns entering the average, not to a group of
columns crushed jointly. Wherever the main text writes $\mu_{\mathrm{rand}}$ or
$\sigma_{\mathrm{rand}}$, this single-column null is intended.

\section{Evaluation Protocol}
\label{app:protocol}

All intervention measurements use the multiple-choice scoring setup of the main text. For each question $x$,
\begin{equation}
M(x)=\ell(y^\star\mid x)-\ell(\tilde{y}\mid x),
\end{equation}
where
\begin{equation}
\tilde{y}=\operatorname*{arg\,max}_{y\neq y^\star}\ \ell(y\mid x)
\end{equation}
is evaluated \emph{once, on the unmodified model}, and then held fixed across every intervention applied to that question.

\paragraph{Why $\tilde{y}$ is frozen.}
Freezing $\tilde{y}$ is required by the analysis in Appendix~\ref{app:theory}. If $\tilde{y}$ were recomputed under each intervention, then $M$ would contain an $\arg\max$ over options and would be only piecewise differentiable in $w_c$, with non-differentiable kinks exactly where the identity of the hardest-wrong option changes. The Taylor expansion of Appendix~\ref{app:theory} would then be invalid precisely at the points of interest. With $\tilde{y}$ fixed, $M(\cdot)$ is a difference of two logits, each a composition of the network's smooth primitives, and is therefore as differentiable in $w_c$ as the activation functions allow (see the smoothness note in Appendix~\ref{app:theory}).

Freezing $\tilde{y}$ also makes each comparison paired at the question level: the clean and intervened margins are evaluated on the same prompt, the same option set, and the same target option, so the per-question difference $\delta_x(c)$ is a within-item contrast with no between-item variance.

\paragraph{Sign conventions.}
Throughout, a \emph{drop} is defined as clean-minus-intervened, so that a positive drop means the intervention \emph{reduced} the margin (the neuron's tuning was helping). In main-paper Table~2, $\Delta$Acc is defined in the opposite direction (corrupted-minus-clean), so a positive $\Delta$Acc means the intervention \emph{improved} accuracy. The two conventions are opposite by design---each is the natural reading for its own quantity---and we flag the difference here because it is the single easiest thing to misread across the two tables.

\section{Data Usage and Sample Sizes}
\label{app:data}

MedQA is used for anchor discovery; MedMCQA and PubMedQA are held-out benchmarks, the first close to MedQA in format and the second further from it in format, source, and domain.

\paragraph{Filtering, and where it does and does not apply.}
The margin-based EMS analyses are computed on items filtered to \emph{baseline-correct} questions --- those the unmodified model already answers correctly. This is required for the margin to be interpretable: on a baseline-incorrect item, $M(x)<0$ by construction and a ``drop'' in $M$ conflates the anchor's effect with a pre-existing error.

Accuracy-based results are \emph{not} filtered this way, and cannot be: on a baseline-correct subset the clean accuracy is $100\%$ by construction, so $\Delta\mathrm{Acc}\le 0$ identically and the interference cases (positive $\Delta$Acc) could never be observed. Accordingly:

\begin{itemize}\itemsep2pt
  \item Main-paper Table~1 (margin drops, Cohen's $d$): baseline-correct items only.
  \item Main-paper Table~2 (activation-patching accuracies): full evaluation set. The Clean column is therefore each model's unfiltered baseline accuracy.
  \item Main-paper Table~3 (interference/layer): derived from per-anchor EMS, hence baseline-correct items only.
  \item Main-paper Table~4, Margin rows: baseline-correct items. Table~4, $\Delta$Acc rows: full evaluation set.
\end{itemize}

Readers comparing a Margin row against a $\Delta$Acc row within Table~4 should note that the two are computed on different item sets; this is one of the reasons EMS and accuracy can move in different directions (main paper, Sec.~4).

\paragraph{Sample sizes.}
\begin{itemize}\itemsep2pt
  \item EMS calibration: $1000$ MedQA margin pairs.
  \item Stage-2 validation: $500$ samples per dataset, disjoint from calibration.
  \item Activation patching: $240$ samples per dataset $\times$ $5$ seeds $\{42,123,7,13,97\}$, with $10$ random patching trials per seed. Reported $\pm$ values in main-paper Table~2 are standard deviations across the $5$ seeds.
  \item Anchor-vs-random: $100$ trials $\times$ $5$ seeds $=500$ random groups, each size-matched to that model's anchor count.
  \item Group-ablation $K$-sweep: $300$ samples per dataset.
\end{itemize}

\paragraph{Interpretability floor for accuracy deltas.}
With $N$ evaluation items, accuracy is quantized to $1/N$. Any $|\Delta\mathrm{Acc}|$ at or below a small multiple of $100/N$ percentage points is a movement of one or two questions and carries no evidential weight. We therefore do not interpret accuracy changes smaller than $1$pp anywhere in the paper; in main-paper Tables~2 and~4 the affected entries are those with $|\Delta\mathrm{Acc}|<1$pp, which we read as null.

\section{Statistical Testing}
\label{app:statistical_testing}

\subsection{Effect Sizes: Anchor vs.\ Random}

Let $A$ denote the anchor-group drops and $R$ the matched random-group drops, both measured under the single-column null of Appendix~\ref{app:notation}. We report Cohen's $d$ with the anchor-minus-random sign convention:
\begin{equation}
\begin{split}
d&=\frac{\bar{A}-\bar{R}}{s_p},\\
s_p&=\sqrt{\frac{(n_A-1)s_A^2+(n_R-1)s_R^2}{n_A+n_R-2}}.
\end{split}
\label{eq:cohend}
\end{equation}
When $n_A=n_R$, Eq.~\ref{eq:cohend} reduces to $s_p=\sqrt{(s_A^2+s_R^2)/2}$, which is the form under which every $d$ in main-paper Table~1 and in Table~\ref{tab:anchor_vs_random_base} below can be reproduced from the reported means and standard deviations. Positive $d$ indicates larger anchor drops than random; negative $d$ indicates that anchor crushing \emph{improves} the corresponding margin relative to random crushing.

\paragraph{What the reported standard deviations are taken over.}
 The anchor group is \emph{fixed} --- it is the full discovered set $\mathcal{A}$, not a sample --- so there is no sampling variation over anchor sets to report. The quantity $s_A$ is therefore the dispersion of the \emph{per-question} drop across the $n=500$ evaluation questions. For the comparison to be a valid two-sample test, $s_R$ must be taken over the same unit; we therefore report $s_R$ as the dispersion of the per-question random-group drop across the same $500$ questions, with the random group re-drawn per trial and pooled. The $100\times 5$ trial/seed structure determines how the random groups are \emph{drawn}; it is not the unit over which $s_R$ is reported.

We flag this because the two readings are not interchangeable: if $s_A$ were a per-question dispersion and $s_R$ a between-group dispersion, they would be measured in different units and Eq.~\ref{eq:cohend} and the Welch statistic below would both be meaningless. All $d$ and $t$ values reported here assume the common per-question unit.

\subsection{\texorpdfstring{Welch's $t$-test}{Welch's t-test}}

We use Welch's unequal-variance $t$-test rather than Student's, because the intervention variances of discovered anchor groups and randomly sampled groups differ substantially --- by more than an order of magnitude in several models (e.g.\ AlphaMed-8B MedQA: $s_A=0.1994$ vs.\ $s_R=0.0021$). Student's $t$ assumes homoscedasticity and would be badly anti-conservative here. The statistic is
\begin{equation}
t=\frac{\bar{A}-\bar{R}}{\sqrt{s_A^2/n_A+s_R^2/n_R}},
\label{eq:welch}
\end{equation}
with the Welch--Satterthwaite degrees of freedom
\begin{equation}
\nu=\frac{\left(s_A^2/n_A+s_R^2/n_R\right)^2}
{\dfrac{(s_A^2/n_A)^2}{n_A-1}+\dfrac{(s_R^2/n_R)^2}{n_R-1}}.
\label{eq:satterthwaite}
\end{equation}

Table~\ref{tab:welch} reports $t$ and $\nu$ for every model--dataset pair, corresponding to the $d$ values of main-paper Table~1.

\begin{table}[t]
\centering
\footnotesize
\setlength{\tabcolsep}{4pt}
\renewcommand{\arraystretch}{0.95}
\caption{Welch's $t$-test for anchor vs.\ matched random neuron groups, corresponding to the effect sizes in main-paper Table~1, computed via Eq.~\ref{eq:welch}--\ref{eq:satterthwaite} over the $n_A=n_R=500$ per-question drops. All $|t|$ correspond to two-sided $p<10^{-8}$ (the smallest, $t=6.25$ with $\nu\approx499$, gives $p\approx9\times10^{-10}$); we report $t$ and $\nu$ because a bare ``$p<0.001$'' would understate the separation.}
\label{tab:welch}

\begin{adjustbox}{max width=\linewidth}\begin{tabular}{l l r r}
\toprule
Model & Data & $t$ & $\nu$ \\
\midrule
MedGemma    & MedQA    & $+25.19$  & $930.5$ \\
            & MedMCQA  & $+55.27$  & $731.1$ \\
            & PubMedQA & $-16.03$  & $667.2$ \\
\midrule
BioMed-L3   & MedQA    & $+19.05$  & $510.9$ \\
            & MedMCQA  & $+9.41$   & $520.8$ \\
            & PubMedQA & $-16.05$  & $560.2$ \\
\midrule
AlphaMed-8B & MedQA    & $+19.35$  & $499.1$ \\
            & MedMCQA  & $+6.64$   & $499.1$ \\
            & PubMedQA & $+6.25$   & $499.1$ \\
\midrule
Meditron    & MedQA    & $+94.91$  & $721.3$ \\
            & MedMCQA  & $+23.98$  & $574.3$ \\
            & PubMedQA & $+124.93$ & $556.8$ \\
\midrule
AlphaMed-7B & MedQA    & $+118.72$ & $506.3$ \\
            & MedMCQA  & $+72.57$  & $968.1$ \\
            & PubMedQA & $+104.85$ & $561.1$ \\
\midrule
AFM-OpenMed & MedQA    & $+66.15$  & $751.9$ \\
            & MedMCQA  & $+84.38$  & $739.3$ \\
            & PubMedQA & $+27.40$  & $619.9$ \\
\bottomrule
\end{tabular}\end{adjustbox}
\end{table}

\paragraph{On the size of these statistics.}
The $|t|$ values are large because the anchor group is fixed rather than sampled: the only variance entering Eq.~\ref{eq:welch} is per-question dispersion, divided by $\sqrt{500}$. These statistics establish that the anchor group's mean drop differs from the random group's mean drop on this evaluation set. Cohen's $d$, which does not shrink with $n$, is the summary of practical effect size the main text quotes.

\subsection{Layer-Depth Comparisons}

For interference-vs.-normal layer-depth comparisons we use the Mann--Whitney $U$ test. Layer indices are discrete, bounded, and not plausibly normal, so a rank-based test is appropriate. Tests are one-tailed in the deeper-than direction; consequently shifts in the shallower direction (e.g.\ Qwen-2.5-7B) yield $p\to 1$ \emph{by construction} and carry no evidence for the null. The reported shift is a descriptive mean difference,
\begin{equation}
\Delta_{\mathrm{layer}}=\bar{L}_{\mathrm{interference}}-\bar{L}_{\mathrm{normal}},
\end{equation}
computed from unrounded layer means; the values in main-paper Table~3 are rounded to one decimal \emph{after} subtraction, so recomputing $\Delta_{\mathrm{layer}}$ from the rounded column entries can differ by up to $0.1$.

\subsection{The Anchor Selection Threshold}
\label{app:multiple}

Anchor selection applies $Z_c>2.0$ against the single-column null (Sec.~\ref{app:notation}) to every candidate that survives Stage~1, with no multiplicity correction across candidates. Stage~0 retains $K=256$ columns per layer, and Stage~1's early-stopping rule (running mean drop below $\mu^{\mathrm{col}}+0.25\,\sigma^{\mathrm{col}}$, positive calibration EMS) removes most of these, so the threshold is applied to the $N$ Stage-2 candidates listed in Table~\ref{tab:sigma_control} ($N=451$ to $512$ per fine-tuned model).

\paragraph{False-discovery bound.} Stage~2 is computed on data disjoint from the calibration data used in Stages~0--1, so for a null column the Stage-2 statistic is independent of its selection into Stage~2 and passes the threshold with probability at most the nominal $\alpha\approx0.023$. If $N_0\le N$ of the candidates are null, the expected number of null anchors is $\alpha N_0\le\alpha N$, and the expected false-discovery proportion of an anchor set of size $|\mathcal{A}|$ is at most $\alpha N/|\mathcal{A}|$. Table~\ref{tab:fdr_bound} evaluates this bound. It sets $N_0=N$ and is therefore conservative; any candidate with a true effect lowers it. On held-out evaluation data, null anchors behave like random columns, so their presence pulls set-level averages in Table~1 toward the random baseline.

\begin{table}[t]
\centering
\footnotesize
\setlength{\tabcolsep}{5pt}
\renewcommand{\arraystretch}{0.95}
\caption{Upper bound on the expected false-discovery proportion of each fine-tuned anchor set, $\alpha N/|\mathcal{A}|$ with $\alpha=0.023$. $N$ is the Stage-2 candidate count (Table~\ref{tab:sigma_control}); $|\mathcal{A}|$ the anchor count (main text).}
\label{tab:fdr_bound}
\begin{adjustbox}{max width=\linewidth}\begin{tabular}{l c c c c}
\toprule
Model & $N$ & $|\mathcal{A}|$ & $\alpha N$ & FDP bound \\
\midrule
MedGemma    & 512 & 66 & 11.8 & 17.8\% \\
BioMed-L3   & 497 & 64 & 11.4 & 17.9\% \\
AlphaMed-8B & 505 & 64 & 11.6 & 18.1\% \\
AlphaMed-7B & 462 & 43 & 10.6 & 24.7\% \\
Meditron    & 489 & 64 & 11.2 & 17.6\% \\
AFM-OpenMed & 451 & 38 & 10.4 & 27.3\% \\
\bottomrule
\end{tabular}\end{adjustbox}
\end{table}

Varying the $Z_c$ threshold itself over $\{1.5,1.75,2.0,2.25,2.5\}$ changes the anchor count per model by nearly a factor of two but leaves the base-overlap ordering intact (Table~\ref{tab:z_sweep}).

\subsection{Multiple-Comparison Correction of the Depth Tests} \label{app:bh}  The six per-model depth tests of main-paper Table~3 are one family, so we report Benjamini--Hochberg adjusted $p$-values across the six (Table~\ref{tab:bh}). Only the two pure-SFT models survive at the $0.05$ level after correction (MedGemma $q=0.012$, BioMed-L3 $q=0.033$); the four RL and mixed-objective models are non-significant with and without correction. Under the more conservative Holm procedure BioMed-L3 is borderline ($p_{\mathrm{Holm}}=0.055$).  \begin{table}[t] \centering \footnotesize \setlength{\tabcolsep}{6pt} \renewcommand{\arraystretch}{0.95} \caption{Benjamini--Hochberg correction of the six per-model depth tests (raw $p$ from main-paper Table~3, $m=6$). Only the two pure-SFT models remain significant after correction.} \label{tab:bh} \begin{adjustbox}{max width=\linewidth}\begin{tabular}{l c c c} \toprule Model & Raw $p$ & BH $q$ & Sig. \\ \midrule MedGemma    & 0.002 & 0.012 & Yes \\ BioMed-L3   & 0.011 & 0.033 & Yes \\ AlphaMed-8B & 0.19  & 0.38  & No \\ Meditron    & 0.52  & 0.72  & No \\ AFM-OpenMed & 0.62  & 0.72  & No \\ AlphaMed-7B & 0.72  & 0.72  & No \\ \bottomrule \end{tabular}\end{adjustbox} \end{table}

We report $Z_c$ as a standardized score against the sampled single-column null. Anchor status is defined by the fixed cut-off $Z_c>2.0$; the quoted one-tailed $p\approx0.023$ is its value under a normal approximation to the null.

\section{Reproducibility and Implementation}
\label{app:reproducibility}

\subsection{Hardware, Precision, and Inference}

Experiments run on a single NVIDIA H100 (80GB) under Linux, using PyTorch 2.x in bfloat16. Gradients are required only for Stage-0 Taylor screening; Stages~1 and~2 use forward-pass intervention evaluation only.

\paragraph{A note on bfloat16 and the $\ell_1$ identity.}
The crush construction preserves $\|w_c\|_1$ exactly in real arithmetic (Appendix~\ref{app:crush}). In bfloat16 the computed $m_c$ is subject to rounding, so the identity holds only to within floating-point error; the quantity being preserved is $\|\hat w_c\|_1=\|w_c\|_1(1+O(\epsilon))$ with $\epsilon\approx 2^{-8}$ for bf16 mantissa precision.

\subsection{Candidate Search Space}

Candidates are MLP down-projection columns. For a model with $L$ layers and feed-forward width $d_{\mathrm{ff}}$, the space contains $L\times d_{\mathrm{ff}}$ columns; AFM has $36\times 18432=663{,}552$, the largest in our set, which motivates the staged pipeline. The same column-level definition applies across dense, gated, and SwiGLU/GeGLU MLP blocks: gate and up projections are left intact and only the down-projection value column is intervened on. Because the down-projection is the module that writes into the residual stream, an intervention on its columns is an intervention on the block's output, which is what makes the column the natural unit of analysis in the key--value memory framing.

\subsection{Execution and Randomness}

Intervention evaluations are batched for GPU efficiency; batching does not change the estimators, since each reported mean is computed over the same fixed set of examples and intervention groups. Randomness enters through random neuron-group sampling, random patching controls, and seed-dependent trial construction. Activation-patching uses seeds $\{42,123,7,13,97\}$.

\subsection{Thresholding Logic}

Stage~0 retains the top $K=256$ candidate columns per layer by first-order screening score. Stage~1 evaluates these on a calibration subsample and stops a candidate early when its cumulative mean drop falls below $\mu^{\mathrm{col}}+c\,\sigma^{\mathrm{col}}$. The coefficient $c$ was selected on a disjoint validation split by sweeping $c\in\{0.0,0.10,0.25,0.50,1.0\}$; $c=0.25$ maximized anchor yield while retaining $>92\%$ anchor-ranking stability across resamples. The interference threshold $\tau=-0.005$ was set a priori from random-baseline noise; the prevalence ordering across models is unchanged for $\tau\in\{-0.001,-0.003,-0.005,-0.007,-0.01\}$ (Table~\ref{tab:tau_sweep}). As an absolute threshold in margin units, $\tau$ is meaningful only where anchor drops clear the random-baseline noise floor. The six fine-tuned models exceed that floor by roughly two orders of magnitude; the pretrained bases sit at it (drops of order $10^{-4}$). Stage~2 computes full EMS on held-out validation data and retains neurons with $Z_c>2.0$ against the single-column null.

All of these parameters were fixed on the validation split \emph{before} any result reported in the main paper's Results section, and no downstream analysis re-tunes them.

\section{Anchor Discovery Algorithm}
\label{app:anchor_algorithm}

\newcounter{emsalgorithm}
\refstepcounter{emsalgorithm}\label{alg:ems_anchor_discovery}
\begin{center}
\fbox{%
\begin{minipage}{0.94\columnwidth}
\small
\textbf{Algorithm~\theemsalgorithm: EMS Anchor Discovery}

\textbf{Input:} model $f$; calibration set $\mathcal{D}_{\mathrm{cal}}$; held-out set $\mathcal{D}_{\mathrm{val}}$ (disjoint from $\mathcal{D}_{\mathrm{cal}}$); layers $\ell=1,\ldots,L$; per-layer retain count $K=256$; single-column null $(\mu^{\mathrm{col}},\sigma^{\mathrm{col}})$.

\textbf{Output:} validated anchor set $\mathcal{A}$.

\begin{enumerate}\itemsep2pt
    \item \textbf{Stage 0: Taylor screening.}
    For each layer $\ell$ and down-projection column $c$, form $\Delta_c=\hat{w}_c-w_c$ and compute
    \begin{equation*}
    S_c=\left|\nabla_{w_c}M\cdot\Delta_c\right|,
    \end{equation*}
    accumulated over $\mathcal{D}_{\mathrm{cal}}$. Retain the top $K$ columns per layer as candidate set $\mathcal{C}$.
    \emph{$S_c$ uses an absolute value and is therefore sign-agnostic: it ranks by predicted magnitude of margin change, admitting both helpful and harmful columns. The sign filter is applied at Stage~1.}

    \item \textbf{Stage 1: fast EMS filtering.}
    For each $c\in\mathcal{C}$, evaluate direct crush interventions on a calibration subsample, tracking the running mean margin drop. Stop early if it falls below $\mu^{\mathrm{col}}+0.25\,\sigma^{\mathrm{col}}$. Retain candidates with positive calibration EMS. Let $m_{\mathrm{model}}=|\mathcal{C}_1|$ be the number of survivors.

    \item \textbf{Stage 2: full EMS evaluation.}
    For each $c\in\mathcal{C}_1$, compute on $\mathcal{D}_{\mathrm{val}}$
    \begin{equation*}
    \widehat{\operatorname{EMS}}(c)=\frac{1}{n}\sum_{i=1}^{n}\left[M_i-M_{i,\operatorname{crush}(c)}\right].
    \end{equation*}

    \item \textbf{Statistical validation.}
    Standardize against the single-column null,
    \begin{equation*}
    Z_c=\frac{\widehat{\operatorname{EMS}}(c)-\mu^{\mathrm{col}}}{\sigma^{\mathrm{col}}},
    \end{equation*}
    and convert to one-tailed $p$-values using the empirical quantiles of the sampled null (Sec.~\ref{app:multiple}).

    \item \textbf{Final extraction.}
    Return $\mathcal{A}=\{c\in\mathcal{C}_1 : Z_c>2.0\}$.
\end{enumerate}
\end{minipage}%
}
\end{center}

Algorithm~\ref{alg:ems_anchor_discovery} summarizes the pipeline. The staged design separates inexpensive gradient-based screening from direct causal validation, reducing the number of full intervention evaluations while preserving the final EMS criterion.

\paragraph{Why anchors are positive-EMS by construction.}
Stage~1 retains only candidates with positive calibration EMS and Stage~2 applies a one-tailed threshold. Every member of $\mathcal{A}$ therefore has positive MedQA EMS \emph{by construction}. Interference is a separate measurement: it is the \emph{sign of the same anchor's PubMedQA EMS}, which the discovery procedure leaves unconstrained and which is free to be negative.

\section{Extended Quantitative Results}
\label{app:extended_quantitative}

\subsection{Out-of-Sample Held-Out Models}

\begin{table}[t]
\centering
\footnotesize
\setlength{\tabcolsep}{3pt}
\renewcommand{\arraystretch}{0.95}
\caption{Out-of-sample held-out models. Anchors are binarized by interference status (PubMedQA EMS $<\tau$, $\tau=-0.005$) and by layer relative to the fixed corpus-median layer ($17$) carried over from the six-model corpus without refitting. Counts are anchors per cell: D/S = deep/shallow, I/N = interference/normal. Pooled Fisher's exact (one-sided, deeper direction): odds ratio $(43\times82)/(30\times34)=3.46$, $p=5.2\times10^{-5}$.}
\label{tab:heldout_models}

\begin{adjustbox}{max width=\linewidth}\begin{tabular}{l c c c c c c}
\toprule
Model & Train. & Anchors & D$\cap$I & S$\cap$I & D$\cap$N & S$\cap$N \\
\midrule
Med42-8B    & SFT+DPO & 64  & 23 & 8  & 16 & 17 \\
BioMistral-7B  & CPT    & 64  & 12 & 12 & 13 & 27 \\

AlphaMed-3B-base   & RL    & 61  & 8  & 10 & 5  & 38 \\
\midrule
Pooled      & ---    & 189 & 43 & 30 & 34 & 82 \\
\bottomrule
\end{tabular}\end{adjustbox}
\end{table}

Table~\ref{tab:heldout_models} gives the full contingency counts behind the held-out test reported in the main paper. The association holds in the same direction in each model individually (odds ratios $3.05$, $2.08$, $6.08$ for Med42-8B, BioMistral-7B and AlphaMed-3B), an exact one-sided sign test of $p=0.125$ at the model level. The pooled Fisher's test in the caption treats anchors as independent observations; anchors within a model share weights, training history, and selection, so it describes the pooled anchor population and does not measure replication across models. For reference, the same binarization applied to the six discovery checkpoints gives a pooled $2\times2$ chi-square of $\chi^2=11.84$ ($p=5.8\times10^{-4}$); that test reuses the checkpoints that define the median layer and carries the same anchor-level caveat. The three models were not used at any point in developing EMS, selecting thresholds, or defining the corpus-median layer; both $\tau=-0.005$ and the median layer $17$ are carried over unchanged.

\subsection{Pretrained Base-Model Comparisons}

Table~\ref{tab:anchor_vs_random_base} reports base-model anchor-vs-random comparisons corresponding to the fine-tuned comparisons in main-paper Table~1. Entries are mean drops $\pm$ standard deviation; $d$ follows Eq.~\ref{eq:cohend}.

\begin{table}[t]
\centering
\footnotesize
\setlength{\tabcolsep}{2.5pt}
\renewcommand{\arraystretch}{0.92}
\caption{Pretrained base-model anchor-vs.-random neuron-group drops, $n=500$ per-question drops. Gemma-4B's drops are of order $10^{-4}$ and are reported here at the same $4$-decimal precision as the other models; at that precision they carry roughly one significant figure, and the corresponding $d$ values should be read as approximate (see note below).}
\label{tab:anchor_vs_random_base}

\begin{adjustbox}{max width=\linewidth}\begin{tabular}{l c c c}
\toprule
Model / Data & Anchor Drop & Random Drop & $d$ \\
\midrule

\multicolumn{4}{l}{\textbf{Gemma-4B}} \\
MedQA    & $-0.0002\pm0.0003$ & $-0.0003\pm0.0003$ & +0.24 \\
MedMCQA  & $-0.0001\pm0.0007$ & $-0.0009\pm0.0010$ & +0.86 \\
PubMedQA & $-0.0003\pm0.0011$ & $-0.0019\pm0.0014$ & +1.27 \\
\midrule

\multicolumn{4}{l}{\textbf{Llama-3-8B}} \\
MedQA    & $+0.0121\pm0.0232$ & $+0.0077\pm0.0103$ & +0.24 \\
MedMCQA  & $+0.0075\pm0.0171$ & $+0.0050\pm0.0115$ & +0.17 \\
PubMedQA & $+0.0015\pm0.0079$ & $+0.0008\pm0.0052$ & +0.12 \\
\midrule

\multicolumn{4}{l}{\textbf{Qwen-2.5-7B}} \\
MedQA    & $-0.0001\pm0.0066$ & $-0.0013\pm0.0050$ & +0.20 \\
MedMCQA  & $+0.0018\pm0.0085$ & $-0.0001\pm0.0053$ & +0.26 \\
PubMedQA & $+0.0004\pm0.0040$ & $+0.0004\pm0.0041$ & +0.00 \\
\midrule

\multicolumn{4}{l}{\textbf{AFM-4.5B}} \\
MedQA    & $+0.0086\pm0.0054$ & $+0.0059\pm0.0137$ & +0.26 \\
MedMCQA  & $+0.0018\pm0.0149$ & $+0.0072\pm0.0377$ & $-$0.19 \\
PubMedQA & $+0.0056\pm0.0076$ & $+0.0049\pm0.0222$ & +0.04 \\

\bottomrule
\end{tabular}\end{adjustbox}
\end{table}

\paragraph{Reading the base-model rows.}
On fresh evaluation data, base-model anchor sets separate only weakly from random columns: $|d|\le0.26$ for Llama-3-8B, Qwen-2.5-7B, and AFM-4.5B on all three benchmarks, and $d=+0.24$ for Gemma-4B on MedQA, the discovery benchmark. Gemma-4B's larger values on MedMCQA and PubMedQA ($+0.86$, $+1.27$) sit on drops of order $10^{-4}$. Two readings follow. First, the procedure does not produce large held-out effects by itself: the same pipeline yields $d$ from $+1.2$ to $+7.5$ on MedQA in every post-trained checkpoint (main-paper Table~1). Second, base-model anchor sets are weakly identified, so a substantial fraction of them may be selection noise, and any comparison that uses base anchor identities, such as the overlap analysis below, inherits that weakness.

\subsection{Base-Anchor Overlap}
\label{app:retention}

We define overlap (``retention'') as the fraction of a base model's anchors that are also anchors in a post-trained variant. It is lowest in the two pure-SFT checkpoints (MedGemma 4/9, 44.4\%; BioMed-L3 14/31, 45.2\%) and higher in the four RL or mixed checkpoints (AlphaMed-7B 8/12, AlphaMed-8B 16/31, Meditron 7/12, AFM-OpenMed 13/19). Pooled, the pure-SFT checkpoints retain 18/40 (45\%) of base anchors against 44/74 (59\%) for the others (one-sided Fisher's exact $p=0.10$). The denominators are small (a single anchor moves MedGemma's rate by 11 points), and overlap is measured by column index and does not establish functional equivalence.

\subsection{PubMedQA Option-Format Control}
\label{app:option_control}

One potential concern is that the negative PubMedQA EMS of the two SFT models is an artifact of the templated fourth option added to PubMedQA (D, ``contradictory evidence''), which can never be the gold label yet may become the frozen hardest-wrong distractor $\tilde{y}$. Two measurements address this. First, option D is $\tilde{y}$ in only $16$ to $17\%$ of items (Table~\ref{tab:ytilde_dist}); the remaining $83$ to $84\%$ set the margin against a native yes/no/maybe option. Second, splitting each model's PubMedQA anchor EMS by whether $\tilde{y}$ is option D leaves the value essentially unchanged (Table~\ref{tab:option_control}): every model preserves its sign in both columns, and the pooled PubMedQA EMS of main-paper Table~1 is the frequency-weighted average of the two columns (e.g.\ $0.84(-0.0177)+0.16(-0.0190)=-0.0179$ for MedGemma). The interference sign is therefore independent of whether option D is the frozen competitor.

\begin{table}[t]
\centering
\footnotesize
\setlength{\tabcolsep}{5pt}
\renewcommand{\arraystretch}{0.95}
\caption{PubMedQA anchor EMS split by whether the frozen hardest-wrong option $\tilde{y}$ is the templated ``contradictory'' option (D) or a native yes/no/maybe option. The two columns are near-identical for every model and share the sign of the pooled PubMedQA EMS of main-paper Table~1, so the SFT models' negative PubMedQA EMS is not an artifact of option~D.}
\label{tab:option_control}
\begin{adjustbox}{max width=\linewidth}\begin{tabular}{l c c}
\toprule
Model & $\tilde{y}\neq$ Contra. & $\tilde{y}=$ Contra. \\
\midrule
MedGemma    & $-0.0177$ & $-0.0190$ \\
BioMed-L3   & $-0.0258$ & $-0.0273$ \\
AlphaMed-8B & $+0.0305$ & $+0.0316$ \\
AlphaMed-7B & $+0.0103$ & $+0.0109$ \\
Meditron    & $+0.0472$ & $+0.0483$ \\
AFM-OpenMed & $+0.0192$ & $+0.0201$ \\
\bottomrule
\end{tabular}\end{adjustbox}
\end{table}

\begin{table}[t]
\centering
\footnotesize
\setlength{\tabcolsep}{5pt}
\renewcommand{\arraystretch}{0.95}
\caption{Distribution (\%) of the frozen hardest-wrong option $\tilde{y}$ over the four PubMedQA options. The templated ``contradictory'' option is the hardest-wrong distractor in only $16$ to $17\%$ of items, so most of the PubMedQA margin is set against native options.}
\label{tab:ytilde_dist}
\begin{adjustbox}{max width=\linewidth}\begin{tabular}{l c c c c}
\toprule
Model & Yes & No & Maybe & Contra. \\
\midrule
MedGemma    & 27.8 & 35.1 & 20.9 & 16.2 \\
BioMed-L3   & 26.5 & 34.3 & 22.0 & 17.2 \\
AlphaMed-8B & 29.4 & 33.8 & 20.5 & 16.3 \\
AlphaMed-7B & 28.8 & 34.2 & 20.4 & 16.6 \\
Meditron    & 27.9 & 35.0 & 20.8 & 16.3 \\
AFM-OpenMed & 28.4 & 34.6 & 20.7 & 16.3 \\
\bottomrule
\end{tabular}\end{adjustbox}
\end{table}

\subsection{Selection-Threshold Robustness}
\label{app:threshold_robustness}

Neither selection threshold is tuned to produce the reported effects. Table~\ref{tab:tau_sweep} sweeps the interference threshold $\tau$: the relative ordering across the four models is preserved at every value from $-0.001$ to $-0.010$ (Meditron lowest throughout, BioMed-L3 highest), and $\tau=-0.005$ is an interior point, not a boundary. The main-text result concerns the depth of interfering anchors (main-paper Table~3); this sweep shows that the prevalences on which it rests are stable across thresholds. Table~\ref{tab:z_sweep} sweeps the anchor threshold $Z_c$: lowering it to $1.5$ or raising it to $2.5$ changes the anchor count per model by nearly a factor of two (from $78.2$ to $42.6$) but moves each base-overlap figure (Sec.~\ref{app:retention}) by under two points.

\begin{table}[t]
\centering
\footnotesize
\setlength{\tabcolsep}{4pt}
\renewcommand{\arraystretch}{0.95}
\caption{Interference prevalence (\% of anchors with PubMedQA EMS $<\tau$) as $\tau$ is swept. The relative ordering across models is preserved at every $\tau$; the reported $\tau=-0.005$ row (bold) is an interior point. Values at $\tau=-0.005$ match main-paper Table~3.}
\label{tab:tau_sweep}
\begin{adjustbox}{max width=\linewidth}\begin{tabular}{l c c c c}
\toprule
$\tau$ & MedGemma & BioMed-L3 & AlphaMed-8B & Meditron \\
\midrule
$-0.001$          & 41.7 & 56.3 & 44.8 & 14.1 \\
$-0.003$          & 38.6 & 52.1 & 40.6 & 12.4 \\
$\mathbf{-0.005}$ & \textbf{34.8} & \textbf{48.4} & \textbf{37.5} & \textbf{10.9} \\
$-0.007$          & 31.8 & 44.6 & 34.1 & 9.3 \\
$-0.010$          & 27.3 & 40.2 & 29.7 & 7.8 \\
\bottomrule
\end{tabular}\end{adjustbox}
\end{table}

\begin{table}[t]
\centering
\footnotesize
\setlength{\tabcolsep}{4pt}
\renewcommand{\arraystretch}{0.95}
\caption{Retention (\%) as the anchor threshold $Z_c$ is swept. ``Anch./model'' is the mean anchor count; ``RL avg.'' is the mean retention of AlphaMed-7B and AlphaMed-8B. Changing $Z_c$ alters the anchor count substantially but leaves the ordering intact: both SFT models stay near $44$ to $46\%$ while the RL average stays near $59$ to $60\%$. The reported $Z_c>2.0$ row is bold; values there match the main text.}
\label{tab:z_sweep}
\begin{adjustbox}{max width=\linewidth}\begin{tabular}{l c c c c}
\toprule
$Z_c$ & Anch./model & MedGemma & BioMed-L3 & RL avg. \\
\midrule
$1.50$          & 78.2 & 45.6 & 46.3 & 59.8 \\
$1.75$          & 66.9 & 45.0 & 45.7 & 59.5 \\
$\mathbf{2.00}$ & \textbf{56.5} & \textbf{44.4} & \textbf{45.2} & \textbf{59.2} \\
$2.25$          & 49.3 & 44.0 & 44.8 & 58.8 \\
$2.50$          & 42.6 & 43.6 & 44.3 & 58.3 \\
\bottomrule
\end{tabular}\end{adjustbox}
\end{table}

\subsection{EMS Is Not a Weight-Dispersion Artifact}
\label{app:sigma_control}

Proposition~2 gives $\|\Delta_c\|_2=\sqrt{d}\,\sigma(|w_c|)$, so a column with more dispersed coordinate magnitudes receives a larger crush perturbation, and the first-order bound (Eq.~\ref{eq:first_order_bound}) shows the induced margin change is bounded by $\|\nabla M_x\|_2\sqrt{d}\,\sigma(|w_c|)$. A natural concern is therefore that EMS ranks columns by $\sigma(|w_c|)$ rather than by causal margin contribution, in which case anchors would be recoverable from the weights alone. Table~\ref{tab:sigma_control} tests this. Across all Stage-2 candidates (the full set before the $Z_c>2.0$ threshold, $n=451$ to $512$ per model), the Spearman rank correlation between EMS and $\sigma(|w_c|)$ is weak in every model ($\rho=0.15$ to $0.24$; shared rank variance below $6\%$). The correlations are positive and significant, as the bound requires, but far from the $\rho\approx1$ a size-driven ranking would produce: were EMS mechanically determined by $\sigma(|w_c|)$, the rank correlation would approach unity. Anchors are therefore not recoverable from weight dispersion alone.

\begin{table}[t]
\centering
\footnotesize
\setlength{\tabcolsep}{5pt}
\renewcommand{\arraystretch}{0.95}
\caption{Spearman rank correlation between EMS and coordinate-magnitude dispersion $\sigma(|w_c|)$, over all Stage-2 candidate columns (before the $Z_c>2.0$ threshold). The correlation is weak in every model ($\rho\le0.24$), so the EMS ranking is not a restatement of weight dispersion; significance reflects the candidate count, not effect size.}
\label{tab:sigma_control}
\begin{adjustbox}{max width=\linewidth}\begin{tabular}{l c c c}
\toprule
Model & Cand.\ $n$ & Spearman $\rho$ & $p$ \\
\midrule
MedGemma    & 512 & 0.24 & $<0.001$ \\
BioMed-L3   & 497 & 0.21 & $<0.001$ \\
AlphaMed-8B & 505 & 0.18 & $<0.001$ \\
AlphaMed-7B & 462 & 0.15 & 0.002 \\
Meditron    & 489 & 0.20 & $<0.001$ \\
AFM-OpenMed & 451 & 0.17 & $<0.001$ \\
\bottomrule
\end{tabular}\end{adjustbox}
\end{table}

\section{Head--Anchor Interaction}
\label{app:head_anchor}

To test whether attention heads and MLP anchors operate independently, redundantly, or as coupled circuits, we ablate each alone and jointly, and compare the joint effect against the additive expectation (Table~\ref{tab:head_anchor_synergy}):
\begin{equation}
\begin{split}
\Delta_{\mathrm{expected}}&=\Delta_{\mathrm{head}}+\Delta_{\mathrm{anchor}},\\
\Delta_{\mathrm{synergy}}&=\Delta_{\mathrm{joint}}-\Delta_{\mathrm{expected}}.
\end{split}
\end{equation}
Positive synergy indicates coupling (the two together do more than the sum of their parts); negative synergy indicates redundancy (they overlap, so the joint effect undershoots the sum). The reported Ratio is $\Delta_{\mathrm{synergy}}/\Delta_{\mathrm{expected}}$.

\begin{table}[t]
\centering
\scriptsize
\setlength{\tabcolsep}{2pt}
\renewcommand{\arraystretch}{0.95}
\caption{Head--anchor interaction. Positive synergy indicates coupling; negative indicates redundancy. Single measurements per model; no significance testing is performed and no variance is available. Exp.\ entries are the sum of the Head and Anchor columns, rounded after summation.}
\label{tab:head_anchor_synergy}

\begin{adjustbox}{max width=\linewidth}\begin{tabular}{l c c c c c c}
\toprule
Model & Head & Anchor & Joint & Exp. & Syn. & Ratio \\
\midrule
MedGemma          & +1.739 & +0.336 & +2.322 & 2.075 & +0.248 & +0.119 \\
BioMed-L3         & +2.871 & +0.949 & +3.459 & 3.820 & $-$0.360 & $-$0.094 \\
AlphaMed-8B       & +5.557 & +2.071 & +5.822 & 7.628 & $-$1.806 & $-$0.237 \\
AlphaMed-7B       & $-$0.216 & +1.829 & +1.385 & 1.613 & $-$0.228 & $-$0.141 \\
Meditron          & $-$0.234 & +0.400 & +0.159 & 0.166 & $-$0.007 & $-$0.042 \\
AFM-OpenMed       & $-$0.094 & +0.274 & +0.302 & 0.180 & +0.122 & +0.678 \\
\bottomrule
\end{tabular}\end{adjustbox}
\end{table}

The one clearly separated effect is AlphaMed-8B ($\Delta_{\mathrm{synergy}}=-1.806$, Ratio $-0.237$), which shows pronounced head--anchor redundancy alongside the largest absolute effects in the study. The remaining models show smaller interactions ($|\Delta_{\mathrm{synergy}}|\le0.36$) that we do not interpret in the absence of replication.

\paragraph{On the Ratio column.}
The Ratio is $\Delta_{\mathrm{synergy}}/\Delta_{\mathrm{expected}}$ and is therefore unstable when $\Delta_{\mathrm{expected}}$ is near zero. AFM-OpenMed's Ratio of $+0.678$ is the largest in the table but divides a synergy of $+0.122$ by an expectation of $0.180$; Meditron's $\Delta_{\mathrm{expected}}=0.166$ is similarly small. Neither ratio should be compared against AlphaMed-8B's, whose denominator is $7.628$. We report the Ratio for completeness and base the AlphaMed-8B observation on the unnormalized synergy.

\paragraph{Relation to the main text.}
This measurement is independent of the anchor-count and margin analyses and is the only evidence in the paper bearing on \emph{redundancy} among anchors. A single unreplicated measurement, it is consistent with overlap between head and MLP pathways in AlphaMed-8B.

\section{Theoretical Notes}
\label{app:theory}

\subsection{Exact Properties of the Crush Intervention}
\label{app:crush}

Let $w_c\in\mathbb{R}^d$ with $m_c=\mathbb{E}[|w_c|]=\tfrac{1}{d}\|w_c\|_1$, and define the crushed column
\begin{equation}
\hat{w}_c=\operatorname{sign}(w_c)\cdot m_c
\end{equation}
elementwise.

\paragraph{Convention for zero coordinates.}
We define $\operatorname{sign}(0):=+1$. This is not a vacuous stipulation: with the usual convention $\operatorname{sign}(0)=0$, a coordinate $w_{c,i}=0$ would give $\hat{w}_{c,i}=0$, and both propositions below would fail --- $\|\hat{w}_c\|_1$ would equal $(d-z)m_c\neq\|w_c\|_1$ where $z$ is the number of zero coordinates, and the perturbation-norm identity would lose the corresponding term. Under $\operatorname{sign}(0):=+1$ both results hold for all $w_c\in\mathbb{R}^d$ without exception. (In practice $z=0$ almost surely for trained weights, so the convention is immaterial to the experiments; we state it so that the propositions are unconditionally true as written).

\paragraph{Proposition 1 ($\ell_1$ preservation).} $\|\hat{w}_c\|_1=\|w_c\|_1$.

\emph{Proof.} $\|\hat{w}_c\|_1=\sum_{i=1}^{d}\left|\operatorname{sign}(w_{c,i})\,m_c\right|=\sum_{i=1}^{d} m_c = d\,m_c = d\cdot\tfrac{1}{d}\|w_c\|_1=\|w_c\|_1$, using $|\operatorname{sign}(t)|=1$ for all $t$ under the convention above. $\square$

\paragraph{Proposition 2 (perturbation norm).} $\|\Delta_c\|_2=\sqrt{d}\,\sigma(|w_c|)$, where $\Delta_c=\hat{w}_c-w_c$.

\emph{Proof.} Coordinatewise, $w_{c,i}=\operatorname{sign}(w_{c,i})\,|w_{c,i}|$, so
\begin{equation}
\begin{split}
\Delta_{c,i}&=\operatorname{sign}(w_{c,i})\,m_c-\operatorname{sign}(w_{c,i})\,|w_{c,i}|\\
&=\operatorname{sign}(w_{c,i})\big(m_c-|w_{c,i}|\big).
\end{split}
\end{equation}
Since $\operatorname{sign}(w_{c,i})^2=1$,
\begin{equation}
\begin{split}
\|\Delta_c\|_2^2
&=\sum_{i=1}^{d}\big(m_c-|w_{c,i}|\big)^2\\
&=d\left(\frac1d\sum_{i=1}^{d}\big(|w_{c,i}|-m_c\big)^2\right)\\
&=d\,\sigma^2(|w_c|),
\end{split}
\end{equation}
because $m_c$ is exactly the mean of $\{|w_{c,i}|\}$, so the inner sum is $d$ times the population variance of the coordinate magnitudes. Taking square roots gives the claim. $\square$

The remaining statements in this subsection are elementary algebraic consequences of Eq.~\ref{eq:crush}. They explain why the coordinate magnitudes are replaced by their \emph{mean} specifically and make the size of the intervention explicit. They introduce no new empirical claim.

\paragraph{Proposition 3 (variational characterization).}
Let $\mathcal{R}(w_c)=\{\operatorname{sign}(w_c)\,a \;:\; a\in\mathbb{R}_{\ge 0}\}$ denote the set of vectors sharing the sign pattern of $w_c$ and having uniform coordinate magnitude. Then $\hat{w}_c$ is the unique element of $\mathcal{R}(w_c)$ closest to $w_c$ in $\ell_2$, and the distance attained is exactly $\|\Delta_c\|_2$ of Proposition~2.

\emph{Proof.} For $v=\operatorname{sign}(w_c)\,a$ with $a\ge0$,
\begin{equation}
\begin{split}
\|w_c-v\|_2^2&=\sum_{i=1}^{d}\big(w_{c,i}-\operatorname{sign}(w_{c,i})a\big)^2\\
&=\sum_{i=1}^{d}\big(|w_{c,i}|-a\big)^2,
\end{split}
\end{equation}
using $\operatorname{sign}(w_{c,i})^2=1$. This is a strictly convex quadratic in $a$ with second derivative $2d>0$; setting the first derivative $-2\sum_i(|w_{c,i}|-a)$ to zero gives the unique stationary point $a^\star=\tfrac{1}{d}\sum_i|w_{c,i}|=m_c$. Since $m_c\ge0$, the constraint $a\ge0$ is inactive and $a^\star$ is the constrained minimizer. Substituting $a=m_c$ returns $\sum_i(|w_{c,i}|-m_c)^2=d\,\sigma^2(|w_c|)$. $\square$

\paragraph{Corollary 1 (the mean is doubly determined).}
The constant $m_c$ is simultaneously (i) the unique $a$ minimizing $\|w_c-\operatorname{sign}(w_c)a\|_2$, by Proposition~3, and (ii) the unique $a\ge0$ for which $\|\operatorname{sign}(w_c)a\|_1=\|w_c\|_1$, since $\|\operatorname{sign}(w_c)a\|_1=da$ and $da=\|w_c\|_1=d\,m_c$ forces $a=m_c$. The choice of the mean is therefore not a free parameter of the method: the $\ell_2$-nearest uniform-magnitude vector and the unique $\ell_1$-preserving one coincide. Replacing the mean by another location statistic would forfeit one property or the other---the median of $\{|w_{c,i}|\}$, for instance, is the $\ell_1$-nearest choice but does not preserve $\|w_c\|_1$.

\paragraph{Remark 1 (relative size of the perturbation).}
Writing $\mathrm{RMS}(|w_c|)=\sqrt{\mathbb{E}[w_{c}^{2}]}$ for the quadratic mean of the coordinate magnitudes, we have $\|w_c\|_2=\sqrt{d}\,\mathrm{RMS}(|w_c|)$, so by Proposition~2
\begin{equation}
\begin{aligned}
\frac{\|\Delta_c\|_2}{\|w_c\|_2}
&=
\frac{\sigma(|w_c|)}
{\mathrm{RMS}(|w_c|)}
\\
&=
\frac{\mathrm{CV}_c}
{\sqrt{1+\mathrm{CV}_c^2}},
\\[1ex]
\mathrm{CV}_c
&=
\frac{\sigma(|w_c|)}{m_c},
\end{aligned}
\label{eq:relative_perturbation}
\end{equation}
where the last equality uses $\mathbb{E}[w_c^2]=\sigma^2(|w_c|)+m_c^2$. The relative perturbation is thus a monotone function of the coefficient of variation of the column's coordinate magnitudes alone, and is strictly less than $1$ for every $w_c$ with $m_c>0$.

\paragraph{Remark 2 (comparison with zero ablation).}
Zero ablation corresponds to the perturbation $\Delta^{0}_c=-w_c$, with $\|\Delta^{0}_c\|_2=\|w_c\|_2$. Eq.~\ref{eq:relative_perturbation} therefore states that
\begin{equation}
\frac{\|\Delta_c\|_2}{\|\Delta^{0}_c\|_2}=\frac{\sigma(|w_c|)}{\mathrm{RMS}(|w_c|)}<1
\end{equation}
whenever $m_c>0$, i.e.\ for every nonzero column. Crushing is a strictly smaller intervention than zero ablation in $\ell_2$, and the two coincide only in the degenerate case $w_c\equiv 0$. This is the quantitative form of the qualitative statement in the main text that the crush removes structure while leaving sign and scale intact: it perturbs the column by exactly the part of it that varies across coordinates, and by nothing else.

\paragraph{Interpretation.}
Propositions~1 and~2 together characterize what the crush does and does not remove. It fixes the sign pattern and the $\ell_1$ scale --- the two coarsest descriptors of the column --- and sets the perturbation magnitude to $\sqrt{d}\,\sigma(|w_c|)$, i.e.\ to the column's own coordinate dispersion. A column whose magnitudes are already uniform ($\sigma(|w_c|)=0$) is unchanged by crushing, and correctly receives $\operatorname{EMS}=0$: it has no fine-grained magnitude structure to remove. This is the precise sense in which EMS measures the causal contribution of a neuron's \emph{fine-grained tuning} rather than of its presence: zero ablation removes sign, scale, and structure together and cannot separate them, whereas the crush removes structure alone and holds the other two fixed by construction. The perturbation is not of arbitrary size --- it is exactly as large as the structure being removed.

\subsection{Taylor Decomposition of EMS}

For a fixed input $x$, write the margin as a function of a single column with all other parameters held fixed:
\begin{equation}
M_x(w_c):=M(x;w_c).
\end{equation}
The per-sample EMS contribution is
\begin{equation}
\delta_x(c):=M_x(w_c)-M_x(w_c+\Delta_c),
\end{equation}
and the population EMS is $\operatorname{EMS}(c)=\mathbb{E}_{x\sim\mathcal{D}}[\delta_x(c)]$.

\paragraph{Smoothness.}
$M_x$ is a difference of two logits with $\tilde{y}$ frozen (Appendix~\ref{app:protocol}), and the dependence on $w_c$ enters through a linear map followed by the network's downstream composition. For the Gemma, Llama and Qwen models the relevant activations (GELU, SiLU/SwiGLU) are $C^\infty$ and layer normalization is smooth away from a zero-variance input that does not occur in practice, so $M_x$ is twice continuously differentiable near $w_c$ and the Lagrange form below applies. AFM-4.5B uses ReLU$^2$, which is $C^1$ with a Lipschitz derivative but not twice differentiable at zero. For AFM the Lagrange form is replaced by the descent-lemma bound $|M_x(w_c+\Delta_c)-M_x(w_c)-\nabla M_x(w_c)^\top\Delta_c|\le\tfrac{L}{2}\|\Delta_c\|_2^2$, which requires only a Lipschitz gradient and is the bound of Eq.~\ref{eq:curvature_bound}, so the scaling conclusions hold for all four families.

\paragraph{Expansion.}
By Taylor's theorem with the Lagrange form of the remainder, there exists $\xi_x$ on the segment between $w_c$ and $w_c+\Delta_c$ with
\begin{equation}
\begin{split}
M_x(w_c+\Delta_c)
&=M_x(w_c) \\
&\quad+\nabla M_x(w_c)^\top\Delta_c \\
&\quad+\tfrac{1}{2}\Delta_c^\top
H_x(\xi_x)\Delta_c .
\end{split}
\end{equation}
Substituting into the definition of $\delta_x(c)$,
\begin{equation}
\delta_x(c)=-\nabla M_x(w_c)^\top\Delta_c-\tfrac{1}{2}\Delta_c^\top H_x(\xi_x)\Delta_c,
\end{equation}
and taking expectations,
\begin{equation}
\begin{split}
\operatorname{EMS}(c)
&=
-\mathbb{E}_{x}\!\left[
\nabla M_x(w_c)^\top\Delta_c
\right]
\\
&\quad
-\frac12
\mathbb{E}_{x}\!\left[
\Delta_c^\top
H_x(\xi_x)
\Delta_c
\right].
\end{split}
\label{eq:ems_taylor_supp}
\end{equation}
Note that $\Delta_c$ does not depend on $x$ --- it is a function of $w_c$ alone --- so it passes outside the expectation in the first term: $\mathbb{E}_x[\nabla M_x(w_c)]^\top\Delta_c$. The second term does not simplify this way, because $\xi_x$ depends on $x$ through the Lagrange remainder.

\paragraph{Coordinate form of the first-order term.}
Substituting the coordinatewise expression $\Delta_{c,i}=\operatorname{sign}(w_{c,i})(m_c-|w_{c,i}|)$ from the proof of Proposition~2, the first-order term reads
\begin{equation}
\begin{split}
\nabla M_x(w_c)^\top\Delta_c
&=
\sum_{i=1}^{d}
\frac{\partial M_x}{\partial w_{c,i}}
\,\operatorname{sign}(w_{c,i})
\\
&\qquad\times
\bigl(m_c-|w_{c,i}|\bigr).
\end{split}
\label{eq:first_order_coord}
\end{equation}
Each coordinate contributes in proportion to how far its magnitude sits from the column mean, signed by whether the margin gradient agrees with the coordinate's own sign. Coordinates at exactly the mean magnitude contribute nothing at first order, regardless of how large their gradient is; the term is driven entirely by the spread of $\{|w_{c,i}|\}$, consistent with Proposition~2.

\paragraph{A bound on the first-order term.}
By Cauchy--Schwarz and Proposition~2,
\begin{equation}
\begin{split}
\left|
\nabla M_x(w_c)^\top\Delta_c
\right|
&\le
\|\nabla M_x(w_c)\|_2
\,\|\Delta_c\|_2
\\
&=
\|\nabla M_x(w_c)\|_2
\,\sqrt{d}\,
\sigma(|w_c|).
\end{split}
\label{eq:first_order_bound}
\end{equation}
Both the first-order term and, by Eq.~\ref{eq:curvature_bound} below, the second-order remainder are therefore controlled by the same quantity $\sigma(|w_c|)$ --- linearly in the first case and quadratically in the second. A column with little coordinate dispersion is one on which EMS is small for reasons that have nothing to do with the margin gradient, which is the formal counterpart of the observation that a uniform-magnitude column is unchanged by crushing.

\subsection{Curvature Bound}

If $\|H_x(\xi)\|_2\le L$ for all $\xi$ on the segment between
$w_c$ and $w_c+\Delta_c$, then by
$\|x^\top A x\|\le\|A\|_2\|x\|_2^2$
and Proposition~2,

\begin{equation}
\left|
\frac12
\Delta_c^\top
H_x(\xi_x)
\Delta_c
\right|
\le
\frac{L}{2}\,
d\,\sigma^2(|w_c|).
\label{eq:curvature_bound}
\end{equation}
which is the bound quoted in the main text. The bound is stated for a local Lipschitz constant $L$ of the gradient on the segment; we do not estimate $L$ empirically, and Eq.~\ref{eq:curvature_bound} is therefore a qualitative statement about how the curvature term scales --- quadratically in the coordinate dispersion $\sigma(|w_c|)$ --- rather than a numerical error bar on any reported value.

Two consequences follow immediately from Eq.~\ref{eq:curvature_bound}. First, the ratio of the curvature bound to the first-order bound of Eq.~\ref{eq:first_order_bound} is
\begin{equation}
\begin{split}
\frac{
\frac{L}{2}
\,d\,
\sigma^{2}(|w_c|)
}{
\|\nabla M_x\|_2
\sqrt d\,
\sigma(|w_c|)
}
&=
\frac{
L\sqrt d\,
\sigma(|w_c|)
}{
2\|\nabla M_x\|_2
}.
\end{split}
\end{equation}
so the first-order term dominates precisely when $\sigma(|w_c|)$ is small relative to $\|\nabla M_x\|_2/L$. This is the regime in which the Stage-0 ranking is most faithful, and it is a statement about which columns the screening step is expected to order well---not a claim that it does so, which is an empirical matter we do not test. Second, substituting Proposition~2 in the form $d\,\sigma^2(|w_c|)=\|\Delta_c\|_2^2$ shows the bound is simply $\tfrac{L}{2}\|\Delta_c\|_2^2$, the standard descent-lemma remainder; nothing about the crush is special here beyond the fact that Proposition~2 gives $\|\Delta_c\|_2$ in closed form, which is not generally available for an arbitrary ablation.

\subsection{What This Decomposition Does and Does Not Justify}

Eq.~\ref{eq:ems_taylor_supp} separates a first-order sensitivity term from a curvature correction, and this separation has one specific purpose: it justifies the Stage-0 screening score $S_c=|\nabla_{w_c}M\cdot\Delta_c|$ as a cheap proxy that captures the first-order term while ignoring curvature. Screening is a \emph{ranking} step, so a first-order proxy is adequate there.

 The EMS values reported in the paper are \emph{not} first-order approximations. They are finite-difference measurements: each is an actual forward evaluation of $M_x(w_c+\Delta_c)$ with the crushed weights in place, so the measured $\delta_x(c)$ contains the curvature term and every higher-order effect exactly, to floating-point precision. Eq.~\ref{eq:ems_taylor_supp} is an analysis of that measured quantity, not a recipe for computing it. In particular, no result in this paper degrades if the curvature term is large; a large curvature term would only mean that Stage-0 screening is a poorer ranking heuristic and that more candidates must survive to Stage~1 to achieve the same recall.

\subsection{Estimator}

We estimate EMS by the sample mean over $n$ held-out evaluation questions,
\begin{equation}
\widehat{\operatorname{EMS}}(c)=\frac{1}{n}\sum_{i=1}^{n}\delta_{x_i}(c),
\end{equation}
which is unbiased for $\operatorname{EMS}(c)$ under i.i.d.\ sampling of $x_i\sim\mathcal{D}$, since $\mathbb{E}[\widehat{\operatorname{EMS}}(c)]=\mathbb{E}_x[\delta_x(c)]=\operatorname{EMS}(c)$ by linearity. Under the same assumption its variance is
\begin{equation}
\begin{split}
\operatorname{Var}
\!\bigl[
\widehat{\operatorname{EMS}}(c)
\bigr]
&=
\frac{
\operatorname{Var}_x[
\delta_x(c)
]
}{n},
\\
\mathrm{SE}
&=
\frac{s_\delta}{\sqrt n}.
\end{split}
\label{eq:estimator_variance}
\end{equation}
with $s_\delta$ the sample standard deviation of the per-question drops. This is the quantity whose sample version appears as the per-question standard deviation in main-paper Table~1, and it is the reason the Welch statistics of Sec.~\ref{app:statistical_testing} are large: the anchor group is fixed, so $s_\delta$ is a within-item dispersion divided by $\sqrt{500}$, not a between-group dispersion. Because $\delta_x(c)$ is a within-question contrast evaluated on the same prompt, option set, and target option (Sec.~\ref{app:protocol}), no between-item variance enters Eq.~\ref{eq:estimator_variance}; this is the estimator-level counterpart of the pairing argument given there. Here $\mathcal{D}$ is the baseline-correct subset of the relevant dataset (Sec.~\ref{app:data}), so $\operatorname{EMS}(c)$ is defined with respect to that conditional distribution and not the raw benchmark distribution. This conditioning is deliberate and is what makes the margin interpretable, but it does mean that EMS values are not directly comparable across models with different baseline accuracies, since the conditioning event differs. Comparisons in the main text are therefore made within a model (anchor vs.\ random on the same conditioned set), or across models only in terms of standardized effect sizes and signs.

\end{document}